\documentclass[letterpaper]{article} % DO NOT CHANGE THIS
\usepackage{aaai2026}  % DO NOT CHANGE THIS
\usepackage{times}  % DO NOT CHANGE THIS
\usepackage{helvet}  % DO NOT CHANGE THIS
\usepackage{courier}  % DO NOT CHANGE THIS
\usepackage[hyphens]{url}  % DO NOT CHANGE THIS
\usepackage{graphicx} % DO NOT CHANGE THIS
\usepackage{natbib}  % DO NOT CHANGE THIS AND DO NOT ADD ANY OPTIONS TO IT
\usepackage{caption} % DO NOT CHANGE THIS AND DO NOT ADD ANY OPTIONS TO IT
\usepackage{algorithm}
\usepackage{algorithmic}

\usepackage{newfloat}
\usepackage{listings}
\DeclareCaptionStyle{ruled}{labelfont=normalfont,labelsep=colon,strut=off} % DO NOT CHANGE THIS
\floatstyle{ruled}
\newfloat{listing}{tb}{lst}{}
\floatname{listing}{Listing}
\usepackage{array}
\usepackage{booktabs}
\usepackage{multirow}
\usepackage{amsmath}
\usepackage{tcolorbox}
\usepackage{arydshln}
\usepackage{amssymb}
\usepackage{adjustbox}

\title{Improving Symbolic Translation of Language Models for Logical Reasoning}
\author {
    Ramya Keerthy Thatikonda\textsuperscript{\rm 1},
    Jiuzhou Han\textsuperscript{\rm 1},
    Wray Buntine\textsuperscript{\rm 2},
    Ehsan Shareghi\textsuperscript{\rm 1}
}
\affiliations {
    \textsuperscript{\rm 1}Department of Data Science \& AI, Monash University\\
    \textsuperscript{\rm 2}College of Engineering and Computer Science, VinUniversity\\
    ramya.thatikonda1@monash.edu
}

\nocopyright

\begin{document}

\maketitle

\begin{abstract}
The use of formal language for deductive logical reasoning aligns well with language models (LMs), where translating natural language (NL) into first-order logic (FOL) and employing an external solver results in a verifiable and therefore reliable reasoning system. However, smaller LMs often struggle with this translation task, frequently producing incorrect symbolic outputs due to formatting and translation errors. Existing approaches typically rely on self-iteration to correct these errors, but such methods depend heavily on the capabilities of the underlying model. To address this, we first categorize common errors and fine-tune smaller LMs using data synthesized by large language models. The evaluation is performed using the defined error categories.
We introduce incremental inference, which divides inference into two stages, predicate generation and FOL translation, providing greater control over model behavior and enhancing generation quality as measured by predicate metrics. This decomposition framework also enables the use of a verification module that targets predicate-arity errors to further improve performance. 
Our study evaluates three families of models across four logical-reasoning datasets. The comprehensive fine-tuning, incremental inference, and verification modules reduce error rates, increase predicate coverage, and improve reasoning performance for smaller LMs, moving us closer to developing reliable and accessible symbolic-reasoning systems.
\end{abstract}

\begin{links}
    \link{Code}{https://github.com/RamyaKeerthy/ProofFOL.git}
\end{links}

\section{Introduction}
The tool-augmented paradigm in logical reasoning tasks with language models (LMs) is to translate statements from natural language (NL) to predicate first-order logic (FOL). These translations are then sent for execution to external solvers such as Z3 \citep{de2008z3} and Prover9 \citep{mccune2005release}, where the tool returns `True', `False', `Uncertain/Unknown' as final output~\citep{ye2024satlm,pan2023logic,olausson2023linc}. Beyond logical reasoning, autoformalisation methods \citep{yang2024formal,wu2022autoformalization} follow a similar symbolic translation pipeline.

\begin{figure*}
    \centering
    \includegraphics[width=0.98\textwidth]{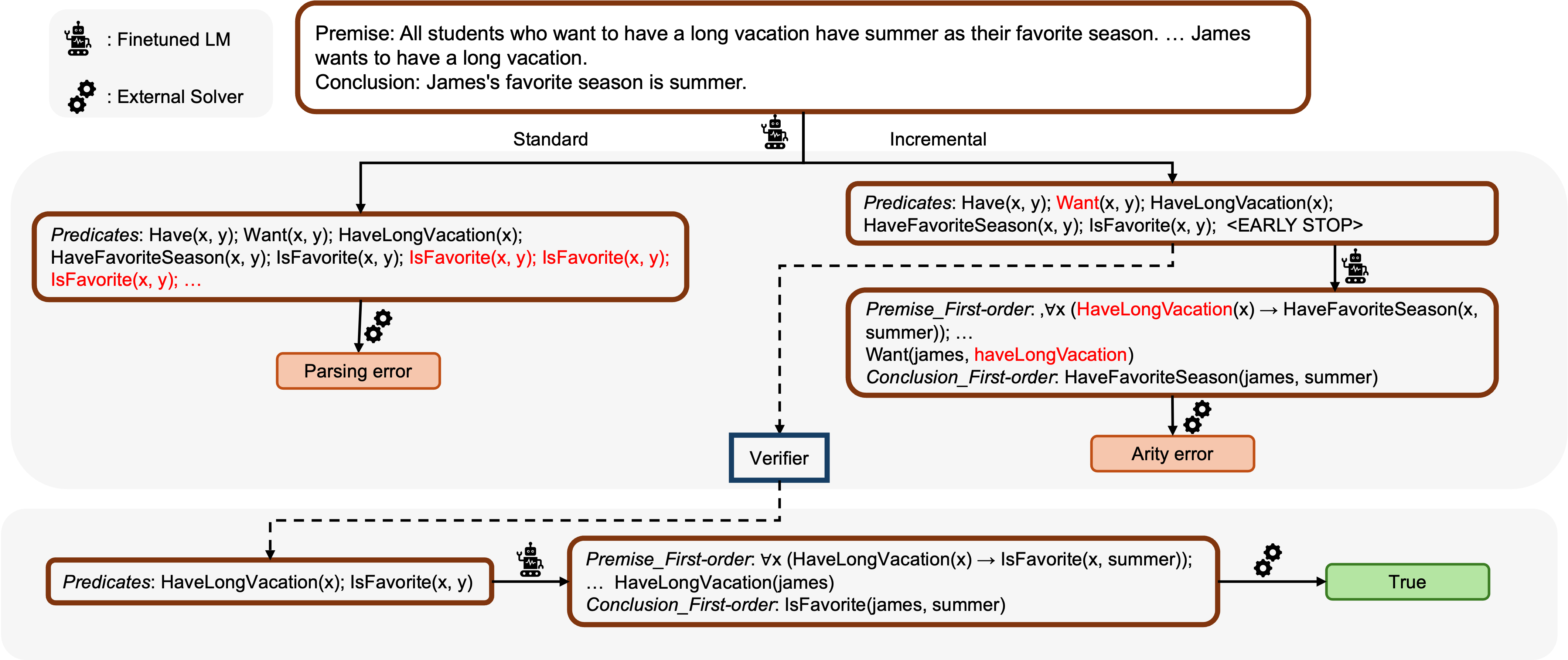}
    \caption{Framework used to improve symbolic translations. Standard inference can still exhibit formatting errors like repetitive loops. Incremental inference mitigates these by stopping predicate generation early and dividing the process into two stages. A verifier helps address arity error with a predicate ``HaveLongVacation" come with arity of 0 and 1 introduced during incremental predicate generation.
    }
    \label{fig:methodology}
\end{figure*}

\citet{yang2023harnessing} highlight systematic errors that even the most advanced LLMs make when translating a \emph{single} NL statement into its corresponding FOL. Realistic logical-reasoning tasks are substantially more challenging, involving \emph{multiple} premise statements followed by a conclusion. These scenarios require consistent NL-to-FOL translations across statements (e.g., in predicate naming or the translation of logical operators). Existing approaches to reducing NL-to-FOL translation errors rely on the LM's ability to understand and self-correct the translation inaccuracies based solely on the error message from an external SMT solver \citep{pan2023logic}. However, both interpreting solver feedback and generating accurate symbolic representations are capabilities more strongly associated with larger models. In contrast, smaller, more accessible language models --- despite their advantages in cost and accessibility for research --- continue to face significant challenges in these tasks.

A key barrier to improving smaller LMs is the scarcity of symbolic data. To bridge this gap, we synthesize additional data for the ProofWriter~\cite{tafjord2020proofwriter} dataset using a large LM, followed by a multi-stage filtering pipeline that includes format correction, syntax validation, and semantic filtering to ensure high translation quality. This synthesized data, combined with the limited existing symbolic dataset \cite{han2022folio}, helps establish a baseline with fewer NL-to-FOL errors (as categorized in our taxonomy) and enables the study of additional frameworks for smaller LMs.

Our fine-tuned models are evaluated under a standard inference setting in which the model is prompted to first generate a list of predicates and then the FOL translations. This design enforces predicate consistency across statements, a requirement for semantic correctness \citep{pan2023logic}. However, under next-token prediction, a single token-level error can cascade throughout the remainder of the generation. As shown in Figure~\ref{fig:methodology}, the inference is stuck in repetitive loops of generating ``IsFavorite(x, y)" \cite{holtzman2019curious}. 
To mitigate this, we introduce \emph{incremental inference}, in which the model first outputs only the set of predicates and then, in a separate step, generates the full FOL statements. This decomposition allows the model to focus on a single subtask at a time, enables early stopping, and ensures well-formatted outputs. While the incremental mode reduces formatting errors, it may still introduce other syntactic issues. The two-stage design therefore facilitates the integration of a plug-and-play verifier module, allowing a lightweight LM to correct predicate-arity errors and further improve translation robustness.

We analyze syntactic errors using our error taxonomy, and we study semantic issues, those that external tools cannot detect, using two predicate-level metrics: \emph{coverage} and \emph{usage}. These metrics quantify the alignment between the predicted predicate set and the predicates appearing in the final FOL statements. Our experiments show that incremental inference substantially improves predicate coverage relative to standard inference.

\begin{table*}[t]
\footnotesize
    \centering
    \renewcommand{\arraystretch}{1.5}
    \resizebox{\textwidth}{!}{%
    \begin{tabular}{p{0.1cm}p{2.5cm}>{\raggedright\arraybackslash}p{11.5cm}}
    \toprule
    \textbf{} & \textbf{Type} & \textbf{Example} \\
    \cline{1-3}
     \multirow{3}{*}{\makebox[0pt][c]{\rotatebox[origin=c]{90}{Parsing}}} &Missing Quantifier & $\text{BerkeleyCollege}(\textcolor{red!60!black}{\underline{x}})$ $\land \text{ResidentialCollegeAt}(\textcolor{red!60!black}{\underline{x}},$ \text{yaleUniversity}) \\
    \cline{2-3}
    &Parenthesis error  & $\text{BeneficialTo}(\text{cherry}, \text{people})$ $\oplus \text{On}(\text{cherry}, \text{warningList})\textcolor{red!60!black}{\underline{)}}$ $\to \neg \text{RedFruit}(\text{cherry})$ \\
    \cline{2-3}
    &Completion error  & $\forall x (\text{Athlete}(x)$ $\to$$ \neg \text{NeverExercises}(x))$ \textcolor{red!60!black}{\underline{\text{Never: does not exist a time}}} \\
    \midrule
    \multirow{2}{*}{\makebox[0pt][c]{\rotatebox[origin=c]{90}{Type}}}&Quantifier location & $\textcolor{red!60!black}{\underline{\exists y}} (\text{Own}(\text{emily}, y)$ $\land \text{Roommate}(y)) \to$ $\textcolor{red!60!black}{\underline{\exists y}} (\text{Own}(\text{emily}, y)$ $\land \text{LiveIn}(\text{emily}, \text{apartment}))$ \\
    \cline{2-3}
    &Missing variable  & $\forall x \exists y (\text{\textcolor{red!60!black}{\underline{In(indonesia)}}} \land \text{Prosecutor}(x) \land \text{SpecialCrime}(y) \rightarrow \text{InvestigatePersonally}(x, y))$ \\ 
    \midrule
    \multirow{2}{*}{\makebox[0pt][c]{\rotatebox[origin=c]{90}{Token}}}& Special $ $    token  & Endowment(yale, \textcolor{red!60!black}{\underline{42.3}} billion) \\ 
    \cline{2-3}
    &Unknown Operator & \begin{tabular}[c]{@{}l@{}}$\forall x (\text{Rating}(x, y) \land y \textcolor{red!60!black}{\underline{>}} 4$$ \rightarrow \text{Listed}(x))$\end{tabular} \\ 
    \midrule
     \multirow{2}{*}{\makebox[0pt][c]{\rotatebox[origin=c]{90}{Predicate}}}& Arity Mismatch &  Sees(Tiger, Mouse); $\forall x$(((Visits(x, Rabbit)) $\land$ \textcolor{red!60!black}{\underline{(Sees(Mouse)))}} $\rightarrow$ (Visits(x, Tiger))) \\
    \cline{2-3}
    &Subject-Predicate Conflict  & $\textcolor{red!60!black}{\underline{\text{Platypus}(\text{platypus})}}$ $\land \neg \text{Teeth}(\text{platypus})$ $\land \text{Mammal}(\text{platypus})$ \\
    \bottomrule
    \end{tabular}}
    \caption{Syntax Errors in First-Order Logic translations. This table categorizes errors by their cause, with the \textit{\textcolor{red!60!black}{\underline{underlined}}} text highlighting the specific cause or location of each error.}
    \label{tab: syntax}
\end{table*}

\section{Related Work}
\paragraph{LMs for Symbolic Translation.}
Formal language translations by large language models (LLMs) was initially attempted by \citet{nye2021improving}, with an intent to emphasize the importance of dual-process theory for logical reasoning tasks. Subsequent work shifted the reasoning burden to theorem provers, with LLMs generating logic programs for deductive reasoning \citep{matthew-lam-etal-2024-closer,pan2023logic, ye2024satlm, olausson2023linc} and inference tasks \citep{liu2025neuro,tarau2025llm}. Research in this direction primarily differs in the choice of formal language (influenced by the underlying theorem prover) and in the verification or correction mechanisms used to address translation errors. Other methods \citep{xu2024faithful} focus on adapting symbolic structures within reasoning paradigms such as chain-of-thought \cite{wei2022chain}, improving reasoning fidelity without employing external provers. However, these approaches often rely on large, expensive, and proprietary models, limiting the deployment of symbolic frameworks in safety-critical or resource-constrained settings.

Complementary to work in formal logic, \citet{yang2023harnessing} applied supervised fine-tuning to LLaMA to improve natural-language to first-order-logic translations at the sentence level. In contrast, our research shifts the focus to building a complete translation system capable of handling multi-statement inputs using accessible and inexpensive small LMs, accompanied by a systematic analysis of translation errors.

\paragraph{Logical Reasoning}
Logical reasoning is a critical capability for evaluating the inference performance of large language models, encompassing three major types: deductive, inductive, and abductive reasoning \citep{xu2025large, liu2025logical}. Our work focuses specifically on deductive reasoning, which requires deriving a logically valid conclusion from a given set of premises. Several datasets support research in this area, including FOLIO \citep{han2022folio}, ProofWriter \citep{tafjord2020proofwriter}, ProntoQA \citep{prontoqa}, ProverQA \citep{qi2025large}, and JustLogic \citep{chen2025justlogic}. Among these, FOLIO is notable for providing complete symbolic translations with human-annotated FOL sentences, but its utility is limited by its relatively small size. As smaller models often struggle to produce meaningful outputs on complex reasoning tasks, we use both existing symbolic datasets and synthetically generated data to supervise-finetune these models and instill the desired behavior \citep{taori_alpaca_2023, ouyang2022training}.

\paragraph{Knowledge Distillation}
Leveraging data generated by larger, more capable models to train smaller models has been widely studied. Early work in this direction focuses on synthetic data generation, where outputs from capable models are used to train instruction-following models \cite{taori_alpaca_2023, wang2023self}. With the growing performance gap between large proprietary models and open-source alternatives, knowledge distillation has gained popularity as a means of transferring capabilities from larger models to more accessible ones \cite{xu2024survey}. Other studies distill rationales from large language models and fine-tune smaller models to enhance their reasoning abilities \cite{hsieh2023distilling, ho2023large}. However, these approaches primarily evaluate effectiveness through downstream performance gains after fine-tuning, offering limited insight into data quality. In contrast, we propose a tool-based symbolic data generation framework that enables external verification, ensuring the reliability and correctness of the data used for fine-tuning.

\paragraph{Structured Inference with LLMs}
The use of multiple prompts distributed across several steps to solve a single problem has been widely studied in LLM inference across domains \cite{yao2022react}. However, these multi-step inferences are typically prompt-induced at each stage. Related work in neurosymbolic and logical reasoning also adopts multi-step prompting. For instance, \citet{ryu273233577divide} first prompts the model to generate step-by-step reasoning and subsequently prompts it again to produce translations; similarly, \citet{tan2025enhancing} first elicits symbolic instances and then performs higher-level inferences over these generated symbols for data synthesis.
Our work differs from this prior line of research in that we rely on a fine-tuned model without requiring additional instructions at inference time. Instead, we use early stopping with smaller LMs to improve output formatting. To the best of our knowledge, ours is the first work to explore this specific form of inference.

\section{NL-to-FOL Translation}
\label{sec:nlfol}
First-order logic (FOL) is a logical framework that uses predicates, quantifiers, variables, and logical connectives, enabling automated reasoning grounded in natural language (NL). Because predicate logic follows strict formal rules \cite{russell2021artificial}, violations of these rules result in syntactic errors, which we discuss in detail in this section.

\subsection{FOL Representation}
We begin by introducing the notation used throughout this work. Let $NL_{P} = \{NL_{P_1}, NL_{P_2}, \dots, NL_{P_n}\}$ denote the set of natural-language premises, and let $NL_C$ denote the natural-language conclusion. Together, they form the natural-language input, $\mathcal{S} \gets \{NL_{P}, NL_C\}$ which is passed to a language model (LM) for symbolic translation. Before generating FOL expressions, we follow the predicate-generation procedure to ensure semantic consistency. This procedure produces a collection of predicates, denoted $\mathcal{P} = \{\, P^{(k)} \mid k = 1, \dots, m \,\}$, where each $P^{(k)}$ is a predicate with $\operatorname{arity}\!\left(P^{(k)}\right) \in \mathbb{N}$, indicating its number of arguments.

The resulting FOL forms of the premises and conclusion are written as $\mathcal{FOL}(\mathcal{S}) = \{\varphi_{P_1}, \varphi_{P_2}, \dots, \varphi_{P_n}, \varphi_{C}\}$, where each $\varphi$ is expressed using the defined predicates. Based on these formulas, the reasoning system produces a final state, assigned one of the following labels: \texttt{True}, \texttt{False}, or \texttt{Uncertain}.

\subsection{Translation Errors}
Large Language Models (LLMs) have demonstrated varying success in translating natural language into formal representations. Among these formalism, NL-FOL translation presents unique challenges involving syntactic and semantic interpretations. In this section, we focus solely on categorizing the syntactic errors observed during LM translations\footnote{Semantic errors such as incorrect quantifiers, operators, or predicates conform to formal syntax but misrepresent the intended meaning. Unlike syntactic errors, they are not automatically detectable through standard parsers or theorem provers. The proxy for detecting semantic errors is the final reasoning accuracy.}. 

Syntactic errors arise from violations of grammatical rules during translation. For example, the rule \textit{``every free variable in a predicate must be bound by a quantifier''} applied to the statement ``Green people are blue'' requires a quantifier `\(\forall\)' for the variable $x$. Missing quantifiers or operators lead to parsing failures, which are often detected by tools such as Prover9 that provide feedback on syntax issues. The most common categories of syntactic errors, summarized in Table~\ref{tab: syntax}, are described below:

\begin{itemize}

    \item \textbf{Parsing errors}: These are caused by structural inconsistencies, such as 
    \begin{itemize}
        \item \emph{Missing quantifier} -- a variable not bound by $\forall$ or $\exists$; 
        \item \emph{Parenthesis error} -- unmatched or missing parentheses render a formula invalid;  
        \item \emph{Completion error} -- incomplete formulas or interruptions by extraneous text.
    \end{itemize}

    \item \textbf{NoneType errors}: These arise from higher-level structural issues, including
    \begin{itemize}
        \item \emph{Quantifier location} -- quantifiers placed incorrectly or redundantly; 
        \item \emph{Missing variable} -- quantifier applied to constant facts with no free variables.
    \end{itemize}

    \item \textbf{Token errors}: These involve invalid lexical elements, such as 
    \begin{itemize}
        \item \emph{Special token} -- unsupported special characters in the input; 
        \item \emph{Unknown operator} -- operators not belonging to the valid set of logical connectives.
    \end{itemize}

    \item \textbf{Predicate errors}: These occur when predicates are used inconsistently, including 
    \begin{itemize}
        \item \emph{Arity mismatch} -- using the same predicate with different numbers of arguments;
        \item \emph{Subject–predicate conflict} -- treating an entity as both a constant and a predicate.
    \end{itemize}
\end{itemize}

For additional details, see Appendix~\ref{app:erroranalysis}.

\section{Methodology}
\subsection{Data synthesis for Finetuning}
\label{sec:datagen}
The alignment of a language model to follow instructions for a specific task can be accomplished by fine-tuning on sufficiently large and representative datasets. Translating natural language into FOL, however, requires capturing formal semantics at the passage level rather than the sentence level, making it challenging for a model to consistently adhere to the required format. To obtain high-quality passage-level translations, we employ a streamlined pipeline for generating FOL data. Our data-generation process builds on ProofWriter, which provides a large number of training instances consisting of multiple premises, a conclusion, and varying reasoning depths. Following \citet{pan2023logic}, we first generate the set of predicates and then derive the corresponding FOL statements for each sentence. We standardize the target formal language to Prover9 and apply it consistently across all translations. Syntax validation is performed using Prover9, which supplies explicit feedback for diverse classes of errors. This synthesized data, combined with existing symbolic datasets, serves as the foundation for training LLMs to produce reliable NL-to-FOL translations. Additional details on the data generation process are provided in Appendix~\ref{appA}.

\paragraph{Supervised Fine-Tuning} 
We apply Supervised Fine-Tuning (SFT) to generate predicates set $\mathcal{P}$, followed by the FOL representations of the premises and conclusion $\mathcal{FOL}(\mathcal{S})$. The model input during training is the concatenation $\mathcal{X} = \mathcal{S} \,\Vert\, \mathcal{P} \,\Vert\, \mathcal{FOL}(\mathcal{S})$, enabling it to learn the structured-form generation pattern. The resulting fine-tuned model is denoted $p_{\theta}$.

\begin{table*}[t]
\normalsize
    \centering
    \resizebox{!}{!}{
    \setlength{\tabcolsep}{4pt}
    \begin{tabular}{llcccccccc}
        \toprule
         &   & \multicolumn{2}{c}{{Gemma-3-4B}} & \multicolumn{2}{c}{{Phi-4-mini}} & \multicolumn{2}{c}{{Qwen-2.5-3B}} & \multicolumn{2}{c}{{Qwen-3-4B}}  \\
        \cmidrule(lr){3-4}\cmidrule(lr){5-6}\cmidrule(lr){7-8}\cmidrule(lr){9-10}
        & {Inference Type} & ExcRate & Acc & ExcRate & Acc & ExcRate & Acc & ExcRate & Acc\\
   \cmidrule(lr){2-2}\cmidrule(lr){3-3}\cmidrule(lr){4-4}\cmidrule(lr){5-5}\cmidrule(lr){6-6}\cmidrule(lr){7-7}\cmidrule(lr){8-8}\cmidrule(lr){9-9}\cmidrule(lr){10-10}
        \parbox[t]{2mm}{\multirow{3}{*}{\rotatebox[origin=c]{90}{PW}}}
        & ICL  & {74.50} & {58.33} & {87.00} & {68.67} & {36.67} & {25.50} & {42.67} & {35.50}   \\
        & SFT\textsubscript{+} & \textbf{98.83}  & \textbf{96.83} & \textbf{100} & \textbf{97.83} & \textbf{96.83}  & \textbf{93.33} & {95.00} & {92.83} \\
        & \hspace{4mm}Increment  & {98.50} & {96.50} & \textbf{100} & \textbf{97.83} & {96.50} & {93.00} & \textbf{96.50} & \textbf{94.17} \\
        \midrule
        \parbox[t]{2mm}{\multirow{3}{*}
        {\rotatebox[origin=c]{90}{FO}}} 
        & ICL  & {58.13} & {33.00} & {45.32} & {28.08} & {27.59} & {19.21} & {6.40} & {3.45}  \\
        & SFT\textsubscript{+}  & {66.00} & {36.95} & {61.08} & \textbf{38.92} & \textbf{57.63} & {30.05} & \textbf{60.10} & \textbf{40.39} \\
        & \hspace{4mm}Increment & \textbf{69.95} & \textbf{40.89} & \textbf{62.07} & \textbf{38.92} & {55.17} & \textbf{30.54} & {58.13} & {38.42} \\
        \midrule
        \multicolumn{10}{c}{\textbf{OOD Datasets}} \\
        \midrule
        % --------------------------------------
        \parbox[t]{2mm}{\multirow{3}{*}{\rotatebox[origin=c]{90}{PrnQA}}} 
        & ICL  & {5.60} & {0.80} & {6.40} & {2.60} & {4.40} & {0.40} & {2.80} & {1.20}  \\
        & SFT\textsubscript{+}  & \textbf{51.80} & {23.20} & {44.60} & {30.60} & \textbf{55.60} & \textbf{21.60} & {62.20} & {51.60}\\
        & \hspace{4mm}Increment  & {50.80} & \textbf{23.60} & \textbf{52.40} & \textbf{34.80} & {54.40} & {19.20} & \textbf{80.60} & \textbf{64.40} \\
        \midrule
        % --------------------------------------
        \parbox[t]{2mm}{\multirow{3}{*}{\rotatebox[origin=c]{90}{PrvQA}}} 
        & ICL  & {28.40}  & {9.40} & {48.40} & {18.20} & {7.60} & {3.00} & {0.00} & {0.00}\\
        & SFT\textsubscript{+}  &  {65.80} & {33.60} & {42.20} & {24.40} &  \textbf{50.40} & \textbf{22.20} & {64.80} & {40.80}\\
        & \hspace{4mm}Increment  & \textbf{69.40} & \textbf{35.60} & \textbf{53.40} & \textbf{31.00} & {50.20} & {20.20} & \textbf{70.40} & \textbf{43.00} \\
        \bottomrule
    \end{tabular}}
    \caption{
        Comparison of models' deductive reasoning accuracy under FOL-based output prediction and the execution rate (\%) of the FOLs by the tool.  ICL denotes in-context learning (5-shots); SFT\textsubscript{+} denotes SFT on symbolic ProofWriter and FOLIO combined; Increment denotes incremental inference on top of the SFT\textsubscript{+} model. The values in \textbf{bold} represent the best performance for each model–dataset pair.
    }
    \label{tab:main-results}
\end{table*}

\subsection{Inference}
In standard inference, the model receives the natural-language statement $\mathcal{S}$ and directly generates both $\mathcal{P}$ and $\mathcal{FOL}(\mathcal{S})$. However, due to the degenerative nature of greedy decoding, where a single error can propagate through the sequence, leading to repetitive or malformed outputs, this approach is brittle. To address this, we introduce \emph{incremental inference}, a two-stage inference procedure.

In the first stage, the model receives the natural-language premises and conclusion and generates a predicate set:

\[
\widehat{\mathcal{P}}
=
\arg\max_{\mathbf{y}}
\, p_{\theta}\!\left(\mathbf{y} \mid \mathcal{S}\right).
\]

In the second stage, the model receives the augmented input $\mathcal{S} \,\Vert\, \widehat{\mathcal{P}}$ and produces the FOL representation:

\[
\widehat{\mathcal{FOL}}(\mathcal{S})
=
\arg\max_{\mathbf{y}}
\, p_{\theta}\!\left(\mathbf{y} \mid \mathcal{S},\, \widehat{\mathcal{P}}\right).
\]

This incremental procedure reduces error propagation and yields more syntactically and semantically coherent FOL outputs. Since the output is decomposed into two stages, any additional latency introduced by incremental inference is attributable to the increased number of output tokens. With an appropriate allocation of output tokens, incremental inference does not incur additional inference time.

\paragraph{Predicate Evaluation}
To analyze the role of predicate generation in NL-to-FOL translation, we use two metrics: \emph{coverage}, analogous to recall, and \emph{usage}, analogous to precision. The model first generates a set of candidate predicates $\mathcal{P_{x}} = \{\, P^{(k)} \mid k = 1, \dots, m \,\}$, followed by the composition of FOL statements. Let the set of predicates that appear in \\$\mathcal{FOL(S)}$ be $\mathcal{P_{y}} = \{\, P^{(k)} \mid k = 1, \dots, n \,\}$

Coverage measures the fraction of predicates used in the FOL statements that were present in the initial generated set, while usage measures the fraction of initially generated predicates that were actually used:  
\[
\begin{aligned}
\mathrm{Coverage}&=\frac{|\mathcal{P_{x}}\cap \mathcal{P_{y}}|}{|\mathcal{P_{y}}|},\\[-2pt]
\mathrm{Usage}&=\frac{|\mathcal{P_{x}}\cap \mathcal{P_{y}}|}{|\mathcal{P_{x}}|}.
\end{aligned}
\]

These predicates from FOL generation can be obtained only if the final output follows a standard format, where it first generates translations for each statement in the premise, and then for the conclusion. We therefore track the \emph{validity} of generations, which measures adherence to this expected format and differs from solver-based execution rates. Our objective is to maximize coverage to ensure accurate predicate dependencies.

\paragraph{Verification}
Because predicate generation is separated from full-statement generation, we additionally explore integrating a predicate verifier before passing the augmented input to the model for FOL generation.

We define a verifier $V$ that checks predicate validity and arity consistency.  
For the generated predicates $\widehat{\mathcal{P}}$ in the first stage of incremental inference, the verifier outputs either the original set or a corrected version:
\[
V\!\left(\widehat{\mathcal{P}}\right) =
\begin{cases}
\widehat{\mathcal{P}}, 
& \text{if predicates satisfy arity consistency}, \\[6pt]
\mathcal{P}^{\mathrm{ver}}, 
& \text{otherwise},
\end{cases}
\]
where $\mathcal{P}^{\mathrm{ver}}$ is the corrected predicate set produced by
the verifier.

The final FOL representation is then obtained in the second stage:

\[
\widehat{\mathcal{FOL}}(\mathcal{S})
=
\arg\max_{\mathbf{y}}
\, p_{\theta}\!\left(\mathbf{y} \mid \mathcal{S},\, V\!\left(\widehat{\mathcal{P}}\right)\right).
\]

\section{Experiments}

We fine-tuned four small-scale models of size 3–4B --- \texttt{Gemma-3-4B-Instruct} \cite{team2025gemma}, \texttt{Phi-4-mini-Instruct} \cite{abouelenin2025phi}, \texttt{Qwen2.5-3B-Instruct} \cite{hui2024qwen2}, and \texttt{Qwen3-4B-Instruct-2507} \cite{qwen3technicalreport} --- on FOL translations from ProofWriter and FOLIO (the latter offering only a limited number of available translations), denoted as SFT\textsubscript{+}. All models were trained for three epochs using supervised fine-tuning with LoRA \citep{hu2022lora}. For inference, we applied 8-bit quantization with a temperature of $0.1$ and a maximum generation limit of $1{,}000$ tokens.

\noindent\paragraph{Benchmarks.} We report the downstream accuracy of the fine-tuned models on four deductive logical reasoning benchmarks: ProofWriter~\cite{tafjord2020proofwriter}, ProntoQA~\cite{prontoqa}, ProverQA~\cite{proverqa}, and FOLIO~\cite{han2022folio}. Notably, ProverQA does not include FOL translations in its training data, and ProntoQA is a test-only benchmark, making inference on these two datasets out-of-distribution (OOD) for our models. For ProofWriter and ProntoQA, we use the test set provided by \citet{pan2023logic}, consisting of 600 and 500 records, respectively. For FOLIO, we used 203 development records for test. For ProverQA, we used 500 records of the `hard' subset of the test set.   
\begin{figure*}
    \centering
    \includegraphics[width=0.98\textwidth]{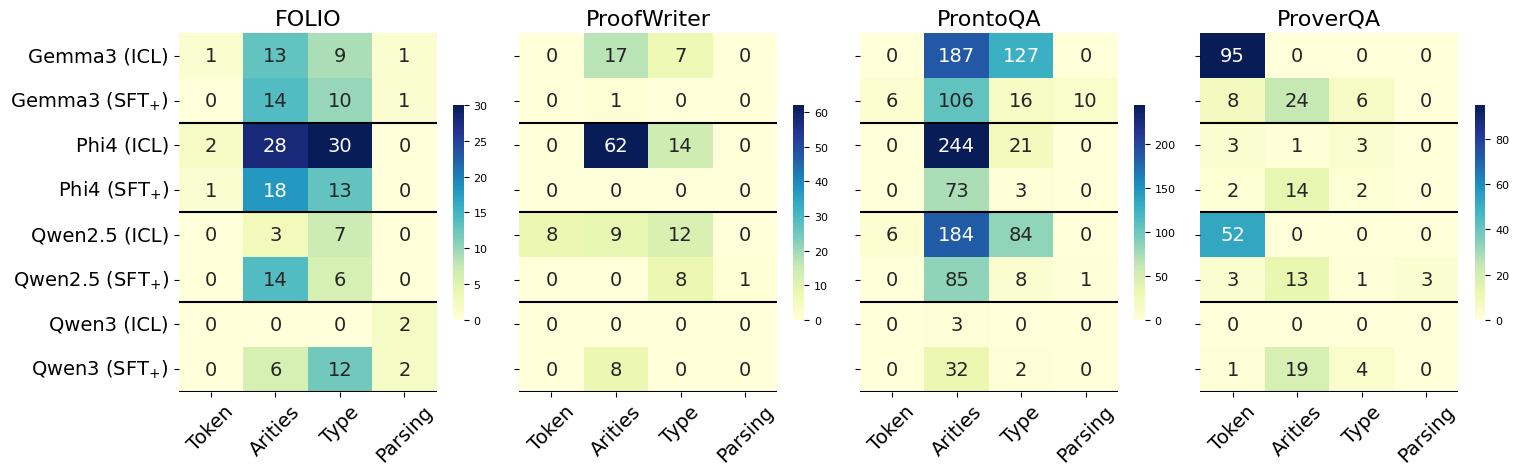}
    \caption{Heatmap of error types counts across datasets, models, and methods. 
    }
    \label{fig:error-count}
\end{figure*}

\begin{figure}[ht]
    \centering
    \includegraphics[width=0.85\linewidth]{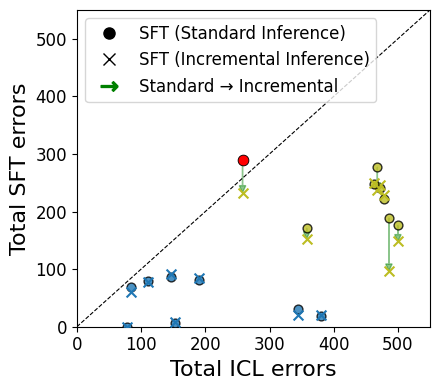}
    \caption{Change in total errors with and without incremental inference across all models and datasets. Blue markers denote in-distribution datasets, while olive green markers denote out-of-distribution (OOD) datasets. Refer to Appendix~\ref{app:errordist} for the numerical values corresponding to this plot.
    }
    \label{fig:movement}
\end{figure}

\begin{table}[t]
\centering
\footnotesize
\setlength{\tabcolsep}{1pt} 
\begin{tabular}{llcccc}
\toprule
 & \textbf{Mode} & \textbf{Gemma-3} & \textbf{Phi-mini} & 
   \textbf{Qwen-2.5} & \textbf{Qwen-3} \\
\midrule
\parbox[t]{2mm}{\multirow{3}{*}{\rotatebox[origin=c]{90}{ProofW}}}
  & Vanilla & 99 / 99 / 100 & 100 / 100 / 100 &  99 / 99 / 100 & 100 / 100 / 100 \\
  & Incr  & 99 / 99 / 100 & 100 / 100 / 100   &  99 / 99 / 100  & 100 / 100 / 100 \\
  & {\small $\Delta$}  &{\small \textcolor{gray}{0 / 0 / 0}} &
                        {\small \textcolor{gray}{0 / 0 / 0}} &
                        {\small \textcolor{gray}{0 / 0 / 0}} &
                        {\small \textcolor{gray}{0 / 0 / 0}}\\
\midrule
\parbox[t]{2mm}{\multirow{3}{*}{\rotatebox[origin=c]{90}{FOLIO}}}
  & Std & 88 / 85 / 93 & 86 / 82 / 91 & 76 / 81 / 97 & 88 / 84 / 96 \\
  & Incr   & 91 / 86 / 96 & 92 / 87 / 97  & 78 / 83 / 97 & 90 / 86 / 100 \\
  & {\small $\Delta$}  &
                        {\small \textcolor{green!60!black}{+3 / +1 / +3}} &
                        {\small \textcolor{green!60!black}{+6 / +5 / +6}} &
                        {\small \textcolor{green!60!black}{+2 / +2 / }} \textcolor{gray}{0} &
                        {\small \textcolor{green!60!black}{+2 / +2 / +4}}\\
\midrule
\parbox[t]{2mm}{\multirow{3}{*}{\rotatebox[origin=c]{90}{Pronto}}}
  & Std  & 80 / 94 / 100 & 49 / 53 / 61 & 72 / 83 / 99 & 69 / 73 / 78 \\
  & Incr  & 81 / 94 / 100 & 53 / 59 / 71  & 72 / 84 / 99 & 75 / 81 / 100 \\
  & {\small $\Delta$}  &
                        {\small \textcolor{green!60!black}{+1 /}} \textcolor{gray}{0 / 0} &
                        {\small \textcolor{green!60!black}{+4 / +6 / +10}}&
                        {\small \textcolor{gray}{0} \textcolor{green!60!black}{/ +1 /}} \textcolor{gray}{0} &     
                        {\small \textcolor{green!60!black}{+6 / +8 / +22}}\\
\midrule
\parbox[t]{2mm}{\multirow{3}{*}{\rotatebox[origin=c]{90}{Prover}}}
  & Std  & 74 / 82 / 91 & 46 / 46 / 53 & 61 / 77 / 92 & 74 / 75 / 87 \\
  & Incr     & 78 / 85 / 96 & 62 / 62 / 76 & 63 / 81 / 97 & 78 / 79 / 95\\
  & {\small $\Delta$}  &
                        {\small \textcolor{green!60!black}{+4 / +3 / +5}} &
                        {\small \textcolor{green!60!black}{+16 / +16 / +23}} &
                        {\small \textcolor{green!60!black}{+2 / +4 / +5}} &
                        {\small \textcolor{green!60!black}{+4 / +4 / +8}}\\
\bottomrule
\end{tabular}
\caption{Predicate analysis across datasets and models. Each entry reports Coverage / Usage / Validity (\%). 
Incr (Incremental) indicates the proposed inference method; $\Delta$ reports (Incr $-$ Vanilla).}
\label{tab:incr-results}
\end{table}

\noindent\paragraph{Evaluation.} We evaluate models using two metrics: execution rate and accuracy. Execution rate is defined as the percentage of valid FOL statements that successfully parse through the tool. We use In-Context Learning (ICL) as a baseline and compare it against SFT for NL-to-FOL translation. In both settings, models generate FOL translations from natural language inputs, and the final logical outcomes are obtained using Prover9. For ProntoQA, which is a test-only dataset, we use demonstrations from ProofWriter for ICL inference, while all other datasets use in-domain few-shot examples. We report results using 5-shot examples for the ICL baseline in our main experiments. Results for different numbers of shots are provided in \S\ref{subsec:shots}.

\section{Results and Discussion}
We evaluate results using accuracy and previously defined error categories. Because the models come from different families and have been trained under distinct settings, we encourage interpreting the results as model-specific and sensitive to the particular data types used.

\subsection{Logical Reasoning with SFT}\label{subsec:error}
Table~\ref{tab:main-results} presents initial results on logical-reasoning datasets using both ICL and SFT, comparing performance across small-scale model families on multiple benchmarks. A consistent and expected trend emerges: fine-tuning on in-domain datasets improves performance. All models show clear gains in both execution rate and accuracy when fine-tuned, correcting the relatively low performance observed under ICL. Notably, fine-tuned models also achieve substantial improvements on OOD datasets, suggesting that training on mixed data is particularly conducive to generalization. This further reinforces the effectiveness of our data synthesis pipeline.

Beyond accuracy, Figure~\ref{fig:error-count} shows the distribution of error categories across all benchmarks. ProntoQA exhibits a large number of arity and type errors under ICL; SFT substantially reduces these errors, with type errors reduced more consistently than arity errors. We attribute this difference to the dataset’s use of fictional abstractions such as “jumpus,” which appear to confuse models during predicate generation.

There are a few cases in which certain error categories increase slightly under SFT, but these do not include formatting errors, which are effectively corrected through fine-tuning. The reduction in formatting issues can, in turn, lead to a slight increase in other types of errors. Further details are provided in Appendix~\ref{app:errordist}. Figure~\ref{fig:movement} provides an overall comparison of SFT error reductions relative to the baseline. We observe decreases across nearly all conditions: ProofWriter shows a 100\% reduction in errors for \texttt{Phi-4-mini-Instruct}; FOLIO shows the largest drop of 57.4\% using \texttt{Qwen3-4B-Instruct-2507}; ProntoQA and ProverQA show 61\% and 64\% reductions, respectively, using \texttt{Qwen3-4B-Instruct-2507}.

One exception, highlighted in red, is a 12\% increase in error rate on ProverQA using Phi-4-mini. In this case, the model struggles to maintain the intended output format due to the large number of premises per query, causing it to lose context during generation. This motivates us to use incremental inference.

\begin{figure}[ht]
    \centering
        \includegraphics[width=\linewidth]{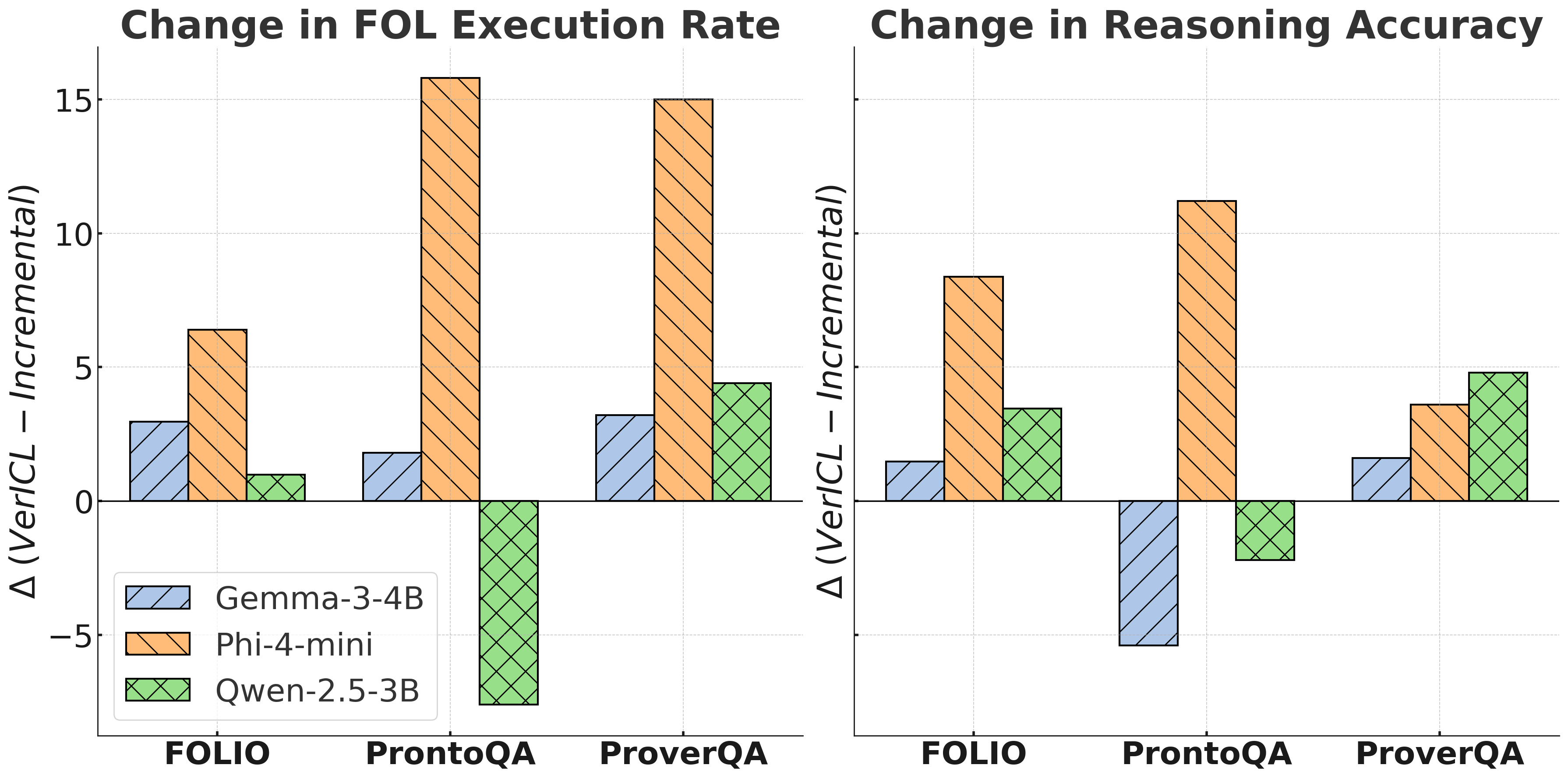}
        \caption{The gain from adding ICL verifier on top of Incremental inference. We use the \texttt{Gemma-3-4B-Instruct} as an ICL verifier with 3-shot examples that include arity perturbations from the respective training data.}
        \label{fig:verify}
\end{figure}

\paragraph{Incremental Inference}
A two-stage generation process enables stricter adherence to the required output format compared to standard inference. Figure~\ref{fig:movement} illustrates this effect: the arrows fall further below the diagonal, indicating roughly a 10\% additional reduction in errors relative to the increase observed under standard inference. Although this improvement is modest, it demonstrates the effectiveness of the method. We anticipate that higher-level or multi-stage forms of incremental inference may yield stronger gains, and we leave such extensions for future work.

\subsection{Predicate Analysis}
Because our earlier evaluation primarily focused on accuracy and tool-detected errors, we introduce an additional analysis targeting the semantic consistency of predicates in the generated FOL translations. In particular, we measure how well the predicates predicted in the first stage correspond to those used in the final FOL statements. Table~\ref{tab:incr-results} reports predicate coverage and usage, along with the validity rate of the generated FOL expressions.

This analysis highlights how incremental inference influences predicate fidelity. We find that incremental inference consistently improves predicate coverage, with a 16\% increase on OOD data, as predicates are explicitly generated in the first stage rather than inferred implicitly during full-sequence decoding. The validity (an adherence to the required format) of FOL statements also improves by a high of 23\% under this setup. ProofWriter, which already achieves high predicate stability, shows no change, indicating that incremental inference does not disrupt performance when models already behave consistently.

\subsection{Arity Verification during Inference}
Given the two-stage generation structure and the explicit predicate decoding enabled by incremental inference, we also explore integrating a verification mechanism directly into the inference pipeline. We use the term \emph{verifier} to refer to a model that takes natural language statements (premises and conclusion) along with LM-generated predicates and outputs either (i) a corrected set of predicates aligned with the NL statements or (ii) the token `correct’ to indicate that the predicates are valid. 

\begin{figure}[ht]
    \centering
    \includegraphics[width=1\linewidth]{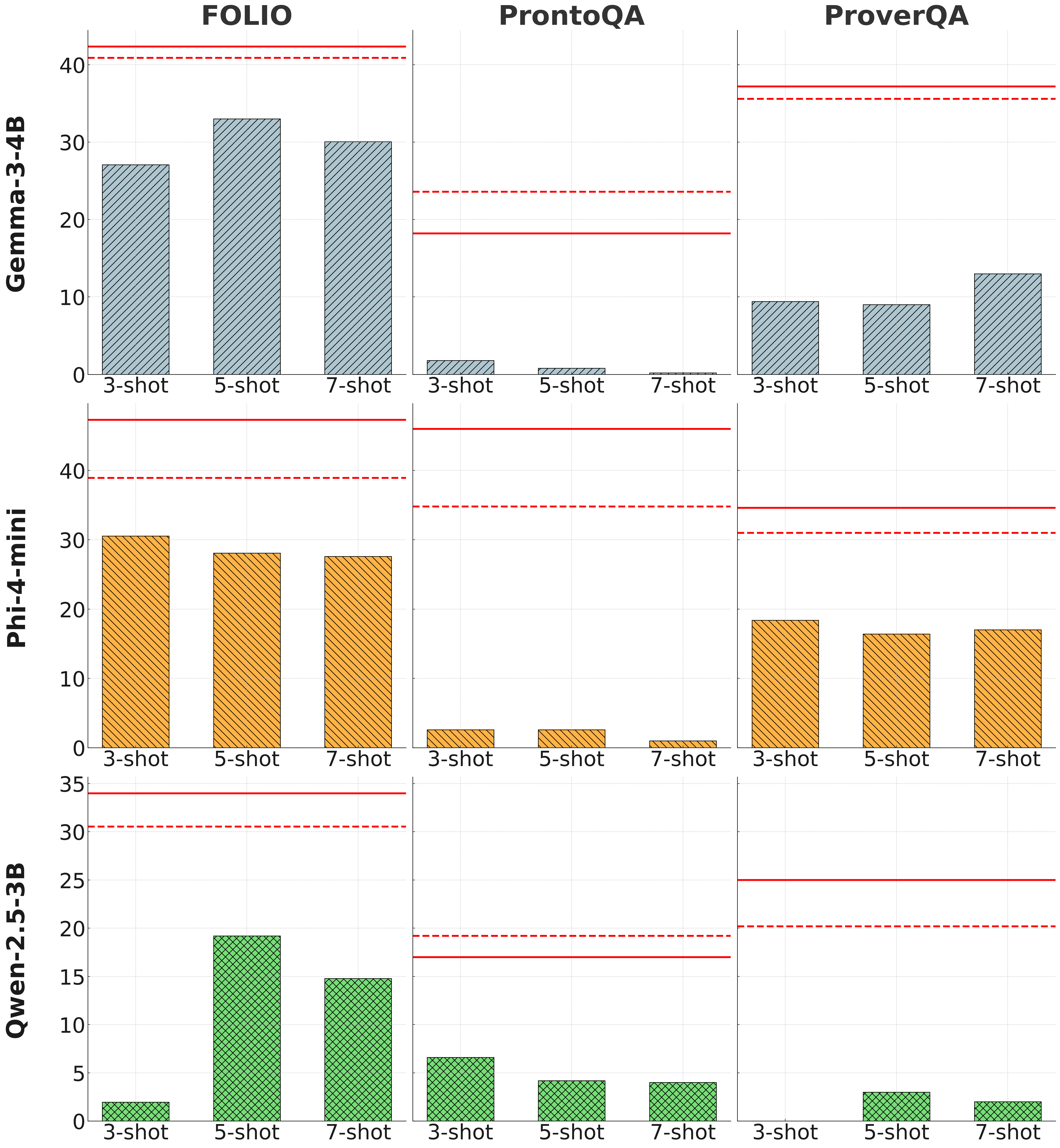}
    \caption{The comparison of reasoning accuracy achieved by using between different number of shots in the translation. The red horizontal lines are for reference: the dashed line reports the results under incremental decoding, and the solid line reports the results with the verification on top.}
    \label{fig:icl}
\end{figure}

We employ \texttt{Gemma-3-4B-Instruct} as the verifier and construct 3-shot demonstrations that include arity errors curated from the training data. Figure~\ref{fig:verify} shows that using a small LM for predicate verification further improves performance over incremental inference alone. As noted earlier, ProntoQA continues to struggle with arity errors due to its semantic abstractions; more tailored in-context examples may help alleviate this gap.
This verification approach can naturally be extended to other types of errors. Full implementation details are provided in Appendix~\ref{app:verifier}.

\subsection{Number of Shots}\label{subsec:shots}
We conduct ICL with varying numbers of few-shot examples randomly sampled from the training sets (3, 5, and 7) to determine the optimal count of demonstrations. The reasoning accuracy results presented in Figure~\ref{fig:icl} show that the performance of smaller language models degrades as more demonstrations are added, particularly for \texttt{Qwen2.5-3B-Instruct} and \texttt{Phi4-mini Instruct}. Although in a few cases accuracy improves with additional few-shots, the gains are negligible compared to the fine-tuned models. The two red horizontal reference lines in each plot indicate the performance achieved by the SFT\textsubscript{+} with incremental inference (dashed line), and the addition of the verifier on top (solid line). See Appendix~\ref{app:ICL} for further ICL results on execution rates as well as the ProofWriter results.

\section{Conclusion}
In this paper, we provide a systematic analysis of common translation errors made by LMs when performing natural language to first-order logic translation. Using small LMs fine-tuned on an expanded set of symbolic data, we compare their error patterns against in-context learning and highlight the limitations of small models under standard inference. We introduce an incremental inference framework and demonstrate that decomposing the generation process improves symbolic outputs due to the structured nature of the task. This method substantially improves predicate generation and reduces formatting errors, including on out-of-distribution datasets. Additionally, we incorporate a lightweight verifier into the first stage of the incremental pipeline, illustrating its potential for further improving translation robustness.
Future work may focus on strengthening the verification module for both predicate and FOL generation, guided by error classification, to further enhance translation accuracy.

\section*{Limitations}
The dataset constructed from ProofWriter inherits limitations in the diversity and complexity of its statements. In addition, the silver data generated using \texttt{GPT-4o}, while carefully curated, may contain minor inconsistencies that we correct to the best of our ability. Fine-tuning performance on logical reasoning tasks is highly dependent on the underlying model, yet direct behavioral comparisons are difficult due to the lack of transparency in pre-training data. For example, Qwen models—particularly the 4B variant—emphasize ``thinking'' during training but perform poorly under in-context learning and tend to generate excessively long reasoning chains during inference, a behavior that becomes more pronounced after fine-tuning and leads to additional computational overhead. 

While the proposed framework ensures coverage between explicitly stated predicates and their symbolic translations, it does not fully address semantic correctness, as symbolic representations may not capture the complete intended meaning of the original natural language statements. Finally, incorporating a verification module introduces additional inference latency, as it adds an extra processing step, although the incremental inference procedure itself does not increase latency.

\bibliography{aaai2026}

@inproceedings{yang2024formal,
  title={Formal mathematical reasoning: A new frontier in ai},
  author={Yang, Kaiyu and Poesia, Gabriel and He, Jingxuan and Li, Wenda and Lauter, Kristin and Chaudhuri, Swarat and Song, Dawn},
  booktitle={arXiv preprint arXiv:2412.16075},
  year={2024}
}

@inproceedings{
proverqa,
title={Large Language Models Meet Symbolic Provers for Logical Reasoning Evaluation},
author={Chengwen Qi and Ren Ma and Bowen Li and He Du and Binyuan Hui and Jinwang Wu and Yuanjun Laili and Conghui He},
booktitle={The Thirteenth International Conference on Learning Representations},
year={2025},
url={https://openreview.net/forum?id=C25SgeXWjE}
}

@article{chen2025justlogic,
  title={Justlogic: A comprehensive benchmark for evaluating deductive reasoning in large language models},
  author={Chen, Michael K and Zhang, Xikun and Tao, Dacheng},
  journal={arXiv preprint arXiv:2501.14851},
  year={2025}
}

@article{qi2025large,
  title={Large language models meet symbolic provers for logical reasoning evaluation},
  author={Qi, Chengwen and Ma, Ren and Li, Bowen and Du, He and Hui, Binyuan and Wu, Jinwang and Laili, Yuanjun and He, Conghui},
  journal={arXiv preprint arXiv:2502.06563},
  year={2025}
}

@article{abouelenin2025phi,
  title={Phi-4-mini technical report: Compact yet powerful multimodal language models via mixture-of-loras},
  author={Abouelenin, Abdelrahman and Ashfaq, Atabak and Atkinson, Adam and Awadalla, Hany and Bach, Nguyen and Bao, Jianmin and Benhaim, Alon and Cai, Martin and Chaudhary, Vishrav and Chen, Congcong and others},
  journal={arXiv preprint arXiv:2503.01743},
  year={2025}
}

@article{team2025gemma,
  author       = {Aishwarya Kamath and
                  Johan Ferret and
                  Shreya Pathak and
                  Nino Vieillard and
                  Ramona Merhej and
                  Sarah Perrin and
                  Tatiana Matejovicova and
                  Alexandre Ram{\'{e}} and
                  Morgane Rivi{\`{e}}re and
                  Louis Rouillard and
                  Thomas Mesnard and
                  Geoffrey Cideron and
                  Jean{-}Bastien Grill and
                  Sabela Ramos and
                  Edouard Yvinec and
                  Michelle Casbon and
                  Etienne Pot and
                  Ivo Penchev and
                  Ga{\"{e}}l Liu and
                  Francesco Visin and
                  Kathleen Kenealy and
                  Lucas Beyer and
                  Xiaohai Zhai and
                  Anton Tsitsulin and
                  R{\'{o}}bert Busa{-}Fekete and
                  Alex Feng and
                  Noveen Sachdeva and
                  Benjamin Coleman and
                  Yi Gao and
                  Basil Mustafa and
                  Iain Barr and
                  Emilio Parisotto and
                  David Tian and
                  Matan Eyal and
                  Colin Cherry and
                  Jan{-}Thorsten Peter and
                  Danila Sinopalnikov and
                  Surya Bhupatiraju and
                  Rishabh Agarwal and
                  Mehran Kazemi and
                  Dan Malkin and
                  Ravin Kumar and
                  David Vilar and
                  Idan Brusilovsky and
                  Jiaming Luo and
                  Andreas Steiner and
                  Abe Friesen and
                  Abhanshu Sharma and
                  Abheesht Sharma and
                  Adi Mayrav Gilady and
                  Adrian Goedeckemeyer and
                  Alaa Saade and
                  Alexander Kolesnikov and
                  Alexei Bendebury and
                  Alvin Abdagic and
                  Amit Vadi and
                  Andr{\'{a}}s Gy{\"{o}}rgy and
                  Andr{\'{e}} Susano Pinto and
                  Anil Das and
                  Ankur Bapna and
                  Antoine Miech and
                  Antoine Yang and
                  Antonia Paterson and
                  Ashish Shenoy and
                  Ayan Chakrabarti and
                  Bilal Piot and
                  Bo Wu and
                  Bobak Shahriari and
                  Bryce Petrini and
                  Charlie Chen and
                  Charline Le Lan and
                  Christopher A. Choquette{-}Choo and
                  CJ Carey and
                  Cormac Brick and
                  Daniel Deutsch and
                  Danielle Eisenbud and
                  Dee Cattle and
                  Derek Cheng and
                  Dimitris Paparas and
                  Divyashree Shivakumar Sreepathihalli and
                  Doug Reid and
                  Dustin Tran and
                  Dustin Zelle and
                  Eric Noland and
                  Erwin Huizenga and
                  Eugene Kharitonov and
                  Frederick Liu and
                  Gagik Amirkhanyan and
                  Glenn Cameron and
                  Hadi Hashemi and
                  Hanna Klimczak{-}Plucinska and
                  Harman Singh and
                  Harsh Mehta and
                  Harshal Tushar Lehri and
                  Hussein Hazimeh and
                  Ian Ballantyne and
                  Idan Szpektor and
                  Ivan Nardini},
  title        = {Gemma 3 Technical Report},
  journal      = {CoRR},
  volume       = {abs/2503.19786},
  year         = {2025},
  url          = {https://doi.org/10.48550/arXiv.2503.19786},
  doi          = {10.48550/ARXIV.2503.19786},
  eprinttype    = {arXiv},
  eprint       = {2503.19786},
  bibsource    = {dblp computer science bibliography, https://dblp.org}
}

@article{hui2024qwen2,
  author       = {Binyuan Hui and
                  Jian Yang and
                  Zeyu Cui and
                  Jiaxi Yang and
                  Dayiheng Liu and
                  Lei Zhang and
                  Tianyu Liu and
                  Jiajun Zhang and
                  Bowen Yu and
                  Kai Dang and
                  An Yang and
                  Rui Men and
                  Fei Huang and
                  Xingzhang Ren and
                  Xuancheng Ren and
                  Jingren Zhou and
                  Junyang Lin},
  title        = {Qwen2.5-Coder Technical Report},
  journal      = {CoRR},
  volume       = {abs/2409.12186},
  year         = {2024},
  url          = {https://doi.org/10.48550/arXiv.2409.12186},
  doi          = {10.48550/ARXIV.2409.12186},
  eprinttype    = {arXiv},
  eprint       = {2409.12186},
  bibsource    = {dblp computer science bibliography, https://dblp.org}
}

@misc{qwen3technicalreport,
      title={Qwen3 Technical Report}, 
      author={Qwen Team},
      year={2025},
      eprint={2505.09388},
      archivePrefix={arXiv},
      primaryClass={cs.CL},
      url={https://arxiv.org/abs/2505.09388}, 
}

@inproceedings{pan2023logic,
    title = "Logic-{LM}: Empowering Large Language Models with Symbolic Solvers for Faithful Logical Reasoning",
    author = "Pan, Liangming  and
      Albalak, Alon  and
      Wang, Xinyi  and
      Wang, William",
    booktitle = "Findings of the Association for Computational Linguistics: EMNLP",
    month = dec,
    year = "2023",
    address = "Singapore",
    url = "https://aclanthology.org/2023.findings-emnlp.248",
    doi = "10.18653/v1/2023.findings-emnlp.248",
    pages = "3806--3824",
}

@inproceedings{wang2023self,
  title={Self-instruct: Aligning language models with self-generated instructions},
  author={Wang, Yizhong and Kordi, Yeganeh and Mishra, Swaroop and Liu, Alisa and Smith, Noah A and Khashabi, Daniel and Hajishirzi, Hannaneh},
  booktitle={Proceedings of the 61st annual meeting of the association for computational linguistics (volume 1: long papers)},
  pages={13484--13508},
  year={2023}
}

@article{xu2024survey,
  title={A survey on knowledge distillation of large language models},
  author={Xu, Xiaohan and Li, Ming and Tao, Chongyang and Shen, Tao and Cheng, Reynold and Li, Jinyang and Xu, Can and Tao, Dacheng and Zhou, Tianyi},
  journal={arXiv preprint arXiv:2402.13116},
  year={2024}
}

@inproceedings{ye2024satlm,
  author       = {Xi Ye and
                  Qiaochu Chen and
                  Isil Dillig and
                  Greg Durrett},
  title        = {SatLM: Satisfiability-Aided Language Models Using Declarative Prompting},
  booktitle    = {Advances in Neural Information Processing Systems 36: Annual Conference
                  on Neural Information Processing Systems, NeurIPS},
  year         = {2023},
  url          = {http://papers.nips.cc/paper\_files/paper/2023/hash/8e9c7d4a48bdac81a58f983a64aaf42b-Abstract-Conference.html},
  bibsource    = {dblp computer science bibliography, https://dblp.org}
}

@inproceedings{yang2023harnessing,
    title = "Harnessing the Power of Large Language Models for Natural Language to First-Order Logic Translation",
    author = "Yang, Yuan  and
      Xiong, Siheng  and
      Payani, Ali  and
      Shareghi, Ehsan  and
      Fekri, Faramarz",
    booktitle = "Proceedings of the 62nd Annual Meeting of the Association for Computational Linguistics (Volume 1: Long Papers)",
    month = aug,
    year = "2024",
    publisher = "Association for Computational Linguistics",
    url = "https://aclanthology.org/2024.acl-long.375",
    pages = "6942--6959",
}

@inproceedings{olausson2023linc,
  author       = {Theo Olausson and
                  Alex Gu and
                  Benjamin Lipkin and
                  Cedegao E. Zhang and
                  Armando Solar{-}Lezama and
                  Joshua B. Tenenbaum and
                  Roger Levy},
  title        = {{LINC:} {A} Neurosymbolic Approach for Logical Reasoning by Combining
                  Language Models with First-Order Logic Provers},
  booktitle    = {Proceedings of the Conference on Empirical Methods in Natural
                  Language Processing, {EMNLP}},
  pages        = {5153--5176},
  year         = {2023},
  url          = {https://doi.org/10.18653/v1/2023.emnlp-main.313},
  doi          = {10.18653/V1/2023.EMNLP-MAIN.313},
  bibsource    = {dblp computer science bibliography, https://dblp.org}
}

@misc{wei2022chain,
  title={Chain-of-thought prompting elicits reasoning in large language models},
  author={Wei, Jason and Wang, Xuezhi and Schuurmans, Dale and Bosma, Maarten and Xia, Fei and Chi, Ed and Le, Quoc V and Zhou, Denny and others},
  journal={Advances in neural information processing systems},
  volume={35},
  pages={24824--24837},
  year={2022}
}

@inproceedings{han2022folio,
    title = "{FOLIO}: Natural Language Reasoning with First-Order Logic",
    author = "Han, Simeng  and
      Schoelkopf, Hailey  and
      Zhao, Yilun  and
      Qi, Zhenting  and
      Riddell, Martin  and
      Zhou, Wenfei  and
      Coady, James  and
      Peng, David  and
      Qiao, Yujie  and
      Benson, Luke  and
      Sun, Lucy  and
      Wardle-Solano, Alexander  and
      Szab{\'o}, Hannah  and
      Zubova, Ekaterina  and
      Burtell, Matthew  and
      Fan, Jonathan  and
      Liu, Yixin  and
      Wong, Brian  and
      Sailor, Malcolm  and
      Ni, Ansong  and
      Nan, Linyong  and
      Kasai, Jungo  and
      Yu, Tao  and
      Zhang, Rui  and
      Fabbri, Alexander  and
      Kryscinski, Wojciech Maciej  and
      Yavuz, Semih  and
      Liu, Ye  and
      Lin, Xi Victoria  and
      Joty, Shafiq  and
      Zhou, Yingbo  and
      Xiong, Caiming  and
      Ying, Rex  and
      Cohan, Arman  and
      Radev, Dragomir",
    booktitle = "Proceedings of the 2024 Conference on Empirical Methods in Natural Language Processing",
    month = nov,
    year = "2024",
    url = "https://aclanthology.org/2024.emnlp-main.1229/",
    doi = "10.18653/v1/2024.emnlp-main.1229",
    pages = "22017--22031"
}

@inproceedings{nye2021improving,
  author       = {Maxwell I. Nye and
                  Michael Henry Tessler and
                  Joshua B. Tenenbaum and
                  Brenden M. Lake},
  title        = {Improving Coherence and Consistency in Neural Sequence Models with
                  Dual-System, Neuro-Symbolic Reasoning},
  booktitle    = {Advances in Neural Information Processing Systems 34: Annual Conference
                  on Neural Information Processing Systems, NeurIPS},
  pages        = {25192--25204},
  year         = {2021},
  url          = {https://proceedings.neurips.cc/paper/2021/hash/d3e2e8f631bd9336ed25b8162aef8782-Abstract.html},
  bibsource    = {dblp computer science bibliography, https://dblp.org}
}

@misc{xu2024faithful,
  title={Faithful Logical Reasoning via Symbolic Chain-of-Thought},
  author={Xu, Jundong and Fei, Hao and Pan, Liangming and Liu, Qian and Lee, Mong-Li and Hsu, Wynne},
  journal={arXiv preprint arXiv:2405.18357},
  year={2024}
}

@inproceedings{wu2022autoformalization,
  title={Autoformalization with large language models},
  author={Wu, Yuhuai and Jiang, Albert Qiaochu and Li, Wenda and Rabe, Markus and Staats, Charles and Jamnik, Mateja and Szegedy, Christian},
  booktitle={Advances in Neural Information Processing Systems},
  volume={35},
  pages={32353--32368},
  year={2022}
}

@misc{ouyang2022training,
  title={Training language models to follow instructions with human feedback},
  author={Ouyang, Long and Wu, Jeffrey and Jiang, Xu and Almeida, Diogo and Wainwright, Carroll and Mishkin, Pamela and Zhang, Chong and Agarwal, Sandhini and Slama, Katarina and Ray, Alex and others},
  journal={Advances in neural information processing systems},
  volume={35},
  pages={27730--27744},
  year={2022}
}

@inproceedings{
hu2022lora,
title={Lo{RA}: Low-Rank Adaptation of Large Language Models},
author={Edward J Hu and yelong shen and Phillip Wallis and Zeyuan Allen-Zhu and Yuanzhi Li and Shean Wang and Lu Wang and Weizhu Chen},
booktitle={International Conference on Learning Representations},
year={2022},
url={https://openreview.net/forum?id=nZeVKeeFYf9}
}

@misc{yao2022react,
  title={React: Synergizing reasoning and acting in language models},
  author={Yao, Shunyu and Zhao, Jeffrey and Yu, Dian and Du, Nan and Shafran, Izhak and Narasimhan, Karthik and Cao, Yuan},
  journal={arXiv preprint arXiv:2210.03629},
  year={2022}
}

@inproceedings{tarau2025llm,
  author       = {Paul Tarau},
  title        = {On LLM-generated Logic Programs and their Inference Execution Methods},
  booktitle    = {Proceedings 40th International Conference on Logic Programming, {ICLP}},
  series       = {{EPTCS}},
  volume       = {416},
  pages        = {1--14},
  year         = {2024},
  month        = feb,
  url          = {https://doi.org/10.4204/EPTCS.416.1},
  doi          = {10.4204/EPTCS.416.1},
  bibsource    = {dblp computer science bibliography, https://dblp.org}
}

@inproceedings{liu2025neuro,
  author       = {Mingyue Liu and
                  Ryo Ueda and
                  Zhen Wan and
                  Katsumi Inoue and
                  Chris G. Willcocks},
  editor       = {Pedro Cabalar and
                  Francesco Fabiano and
                  Martin Gebser and
                  Gopal Gupta and
                  Theresa Swift},
  title        = {Neuro-Symbolic Contrastive Learning for Cross-domain Inference},
  booktitle    = {Proceedings 40th International Conference on Logic Programming, {ICLP}
                  2024},
  series       = {{EPTCS}},
  volume       = {416},
  pages        = {78--94},
  year         = {2024},
  month        = feb,
  url          = {https://doi.org/10.4204/EPTCS.416.6},
  doi          = {10.4204/EPTCS.416.6},
  bibsource    = {dblp computer science bibliography, https://dblp.org}
}

@inproceedings{de2008z3,
  title={Z3: An efficient SMT solver},
  author={De Moura, Leonardo and Bj{\o}rner, Nikolaj},
  booktitle={International conference on Tools and Algorithms for the Construction and Analysis of Systems},
  pages={337--340},
  year={2008},
  organization={Springer}
}

@inproceedings{mccune2005release,
  title={Release of {P}rover9},
  author={McCune, William},
  booktitle={Mile high conference on quasigroups, loops and nonassociative systems, Denver, Colorado},
  year={2005}
}

@inproceedings{tafjord2020proofwriter,
  author       = {Oyvind Tafjord and
                  Bhavana Dalvi and
                  Peter Clark},
  title        = {ProofWriter: Generating Implications, Proofs, and Abductive Statements
                  over Natural Language},
  booktitle    = {Findings of the Association for Computational Linguistics: {ACL/IJCNLP}
                  2021, Online Event, August 1-6, 2021},
  series       = {Findings of {ACL}},
  volume       = {{ACL/IJCNLP} 2021},
  pages        = {3621--3634},
  publisher    = {Association for Computational Linguistics},
  year         = {2021},
  url          = {https://doi.org/10.18653/v1/2021.findings-acl.317},
  doi          = {10.18653/V1/2021.FINDINGS-ACL.317},
  bibsource    = {dblp computer science bibliography, https://dblp.org}
}

@inproceedings{prontoqa,
  author       = {Abulhair Saparov and
                  He He},
  title        = {Language Models Are Greedy Reasoners: {A} Systematic Formal Analysis
                  of Chain-of-Thought},
  booktitle    = {The Eleventh International Conference on Learning Representations,
                  {ICLR} 2023, Kigali, Rwanda, May 1-5, 2023},
  publisher    = {OpenReview.net},
  year         = {2023},
  url          = {https://openreview.net/forum?id=qFVVBzXxR2V},
  bibsource    = {dblp computer science bibliography, https://dblp.org}
}

@inproceedings{ho2023large,
  title={Large language models are reasoning teachers},
  author={Ho, Namgyu and Schmid, Laura and Yun, Se-Young},
  booktitle={Proceedings of the 61st annual meeting of the association for computational linguistics (volume 1: long papers)},
  pages={14852--14882},
  year={2023}
}

@inproceedings{hsieh2023distilling,
  title={Distilling step-by-step! outperforming larger language models with less training data and smaller model sizes},
  author={Hsieh, Cheng-Yu and Li, Chun-Liang and Yeh, Chih-Kuan and Nakhost, Hootan and Fujii, Yasuhisa and Ratner, Alex and Krishna, Ranjay and Lee, Chen-Yu and Pfister, Tomas},
  booktitle={Findings of the Association for Computational Linguistics: ACL 2023},
  pages={8003--8017},
  year={2023}
}

@inproceedings{matthew-lam-etal-2024-closer,
    title = "A Closer Look at Tool-based Logical Reasoning with {LLM}s: The Choice of Tool Matters",
    author = "Matthew Lam, Long Hei  and
      Thatikonda, Ramya Keerthy  and
      Shareghi, Ehsan",
    editor = "Baldwin, Tim  and
      Rodr{\'i}guez M{\'e}ndez, Sergio Jos{\'e}  and
      Kuo, Nicholas",
    booktitle = "Proceedings of the 22nd Annual Workshop of the Australasian Language Technology Association",
    month = dec,
    year = "2024",
    address = "Canberra, Australia",
    publisher = "Association for Computational Linguistics",
    url = "https://aclanthology.org/2024.alta-1.4/",
    pages = "41--63"
}

@book{russell2021artificial,
  author       = {Stuart Russell and
                  Peter Norvig},
  title        = {Artificial Intelligence: {A} Modern Approach (4th Edition)},
  publisher    = {Pearson},
  year         = {2020},
  url          = {http://aima.cs.berkeley.edu/},
  isbn         = {9780134610993},
  bibsource    = {dblp computer science bibliography, https://dblp.org}
}

@inproceedings{holtzman2019curious,
  author       = {Ari Holtzman and
                  Jan Buys and
                  Li Du and
                  Maxwell Forbes and
                  Yejin Choi},
  title        = {The Curious Case of Neural Text Degeneration},
  booktitle    = {8th International Conference on Learning Representations, {ICLR} 2020,
                  Addis Ababa, Ethiopia, April 26-30, 2020},
  publisher    = {OpenReview.net},
  year         = {2020},
  url          = {https://openreview.net/forum?id=rygGQyrFvH},
  bibsource    = {dblp computer science bibliography, https://dblp.org}
}

@article{xu2025large,
  author       = {Fangzhi Xu and
                  Qika Lin and
                  Jiawei Han and
                  Tianzhe Zhao and
                  Jun Liu and
                  Erik Cambria},
  title        = {Are Large Language Models Really Good Logical Reasoners? {A} Comprehensive
                  Evaluation and Beyond},
  journal      = {{IEEE} Trans. Knowl. Data Eng.},
  volume       = {37},
  number       = {4},
  pages        = {1620--1634},
  year         = {2025},
  url          = {https://doi.org/10.1109/TKDE.2025.3536008},
  doi          = {10.1109/TKDE.2025.3536008},
  bibsource    = {dblp computer science bibliography, https://dblp.org}
}

@article{liu2025logical,
  author       = {Hanmeng Liu and
                  Zhizhang Fu and
                  Mengru Ding and
                  Ruoxi Ning and
                  Chaoli Zhang and
                  Xiaozhang Liu and
                  Yue Zhang},
  title        = {Logical Reasoning in Large Language Models: {A} Survey},
  journal      = {CoRR},
  volume       = {abs/2502.09100},
  year         = {2025},
  url          = {https://doi.org/10.48550/arXiv.2502.09100},
  doi          = {10.48550/ARXIV.2502.09100},
  eprinttype    = {arXiv},
  eprint       = {2502.09100},
  bibsource    = {dblp computer science bibliography, https://dblp.org}
}

@misc{taori_alpaca_2023,
  title        = {Alpaca: A Strong, Replicable Instruction-Following Model},
  author       = {Taori, Rohan and Gulrajani, Ishaan and Zhang, Tianyi and Dubois, Yann and Li, Xuechen and Guestrin, Carlos and Liang, Percy and Hashimoto, Tatsunori B.},
  year         = {2023},
  howpublished = {GitHub repository},
  note         = {\url{https://github.com/tatsu-lab/stanford_alpaca}},
  url          = {https://github.com/tatsu-lab/stanford_alpaca}
}

@inproceedings{ryu273233577divide,
  author       = {Hyun Ryu and
                  Gyeongman Kim and
                  Hyemin S. Lee and
                  Eunho Yang},
  title        = {Divide and Translate: Compositional First-Order Logic Translation
                  and Verification for Complex Logical Reasoning},
  booktitle    = {The Thirteenth International Conference on Learning Representations,
                  {ICLR} 2025, Singapore, April 24-28, 2025},
  publisher    = {OpenReview.net},
  year         = {2025},
  url          = {https://openreview.net/forum?id=09FiNmvNMw},
  bibsource    = {dblp computer science bibliography, https://dblp.org}
}

@inproceedings{tan2025enhancing,
  title={Enhancing logical reasoning in language models via symbolically-guided monte carlo process supervision},
  author={Tan, Xingwei and Valentino, Marco and Akhter, Mahmud Elahi and Liakata, Maria and Aletras, Nikolaos},
  booktitle={Proceedings of the 2025 Conference on Empirical Methods in Natural Language Processing},
  pages={31874--31888},
  year={2025}
}

\appendix
\section{Error Categorization.}
\label{app:erroranalysis}
The error categories are derived from the feedback provided by the tool. We categorize these errors based on common patterns observed in language model generations (Table~\ref{tab:error_patterns}). Formatting errors are grouped together as cases where the FOL statements never reach the tool, as these generations have incompatible or invalid FOL formatting. These are usually the ones with no feedback from the tool, but fail at parsing stage.

\begin{table}[ht]
\centering
\begin{tabular}{|l|p{4cm}|}
\hline
\textbf{Error Category} & \textbf{Representative Pattern} \\ \hline
\textbf{NoneType Error} & \textsc{'NoneType' object has no attribute 'rstrip'} \\ \hline
\textbf{Arity Error} & \textsc{The following symbols are used with multiple arities} \\ 
\hline
\textbf{Parsing Error} & \textsc{A term cannot be constructed from the marked string} \\ 
\hline
\textbf{Token Error} & \textsc{Unexpected token:'-'} \\ \hline
\end{tabular}
\caption{Common error categories and representative patterns used to extract LM-generated errors}
\label{tab:error_patterns}
\end{table}

\begin{table*}[t]
    \centering
    \resizebox{!}{!}{
    \setlength{\tabcolsep}{4pt}
    \begin{tabular}{llcccccccc}
        \toprule
         &   & \multicolumn{2}{c}{{Gemma-3-4B}} & \multicolumn{2}{c}{{Phi-4-mini}} & \multicolumn{2}{c}{{Qwen-2.5-3B}} & \multicolumn{2}{c}{{Qwen-3-4B}}    \\
        \cmidrule(lr){3-4}\cmidrule(lr){5-6}\cmidrule(lr){7-8}\cmidrule(lr){9-10}
        & {Type} & ExcRate & Acc & ExcRate & Acc & ExcRate & Acc & ExcRate & Acc \\
   \cmidrule(lr){2-2}\cmidrule(lr){3-3}\cmidrule(lr){4-4}\cmidrule(lr){5-5}\cmidrule(lr){6-6}\cmidrule(lr){7-7}\cmidrule(lr){8-8}\cmidrule(lr){9-9}\cmidrule(lr){10-10}
        \parbox[t]{2mm}{\multirow{3}{*}{\rotatebox[origin=c]{90}{ProofW}}}
        & 3-shot  & {66.33} & {59.68} & {74.50} & {57.67} & \textbf{57.83} & \textbf{39.67} & {1.33} & {0.83} \\
        & 5-shot  & {74.50} & {58.33} & \textbf{87.00} & \textbf{68.67} & {36.67} & {25.50} & {42.67} & {35.50} \\
        & 7-shot  & \textbf{83.00} & \textbf{64.67} & {26.67} & {23.67} & {15.17} & {10.17} & \textbf{62.83} & \textbf{53.83} \\
         \midrule
        \parbox[t]{2mm}{\multirow{3}{*}{\rotatebox[origin=c]{90}{FOLIO}}}
        & 3-shot  & {46.79} & {27.09} & {42.85} & \textbf{30.05} & {22.16} & {12.81} & {1.97} & {1.48} \\
        & 5-shot  & \textbf{58.13} & \textbf{33.00} & \textbf{45.32} & {28.08} & \textbf{27.59} & \textbf{19.21} & \textbf{6.40} & {3.45} \\
        & 7-shot  & {51.72} & {30.05} & {43.84} & {27.59} & {22.17} & {14.78} & \textbf{6.40} & \textbf{3.94} \\
        \midrule
        \parbox[t]{2mm}{\multirow{3}{*}{\rotatebox[origin=c]{90}{Pronto}}}
        & 3-shot  & \textbf{13.60} & \textbf{1.80} & {6.00} & \textbf{2.60} & {6.80} & \textbf{0.60} & {0.00} & {0.00} \\
        & 5-shot  & {5.60} & {0.80} & \textbf{6.40} & \textbf{2.60} & {4.40} & {0.40} & {2.80} & {1.20} \\
        & 7-shot  & {3.00} & {0.20} & {1.6} & {1.00} & \textbf{7.00} & {0.40} & \textbf{9.40} & \textbf{4.20} \\
        \midrule
        \parbox[t]{2mm}{\multirow{3}{*}{\rotatebox[origin=c]{90}{Prover}}}
        & 3-shot  & {30.80} & {9.40} & {47.20} & \textbf{18.40} & {0.20} & {0.20} & {0.00} & {0.00} \\
        & 5-shot  & {28.40} & {9.40} & \textbf{48.40} & {18.20} & \textbf{7.60} & \textbf{3.00} & {0.00} & {0.00} \\
        & 7-shot  & \textbf{35.80} & \textbf{13.00} & {46.20} & {17.00} & {3.40} & {2.00} & {0.00} & {0.00} \\
       
        \bottomrule
    \end{tabular}}
    \caption{
        Performance of smaller language models using in-context learning of varying sizes of demonstrations. 
    }
    \label{tab:icldem}
\end{table*}

\section{Data Synthesis}
\label{appA}
\begin{figure}[ht]
  \centering
  \includegraphics[width=\columnwidth]{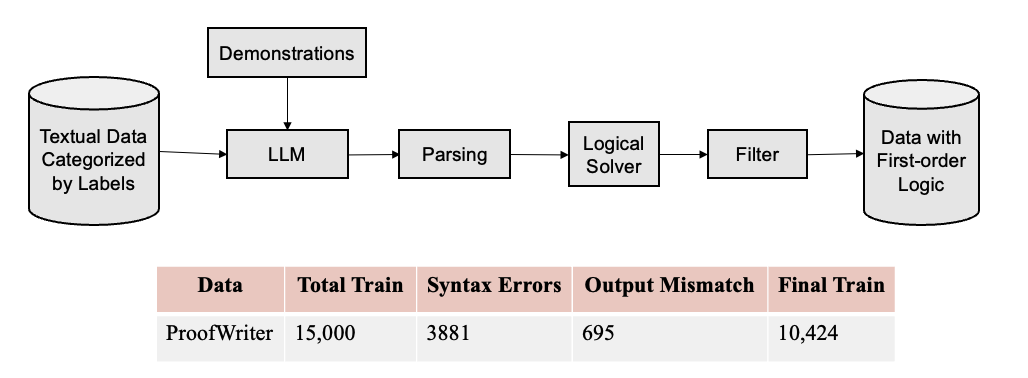}
  \caption{Overview of the Data Generation Pipeline and Key Statistics.}
  \label{fig:gen}
\end{figure}

The data pipeline applied for generating the FOL translations is detailed in Fig \ref{fig:gen}. The train data is $15{,}000$ data points with depth-5 from ProofWriter dataset. The 15k records are sampled randomly ensuring a fair distribution of the labels; True, False, and Uncertain. The LLM here is GPT4o with a output token length of $1{,}000$ for each generation. We ran a small scale experiment through GPT-4o for various few-shot combinations. We selected the most optimal demonstrations with the least format and translation issues. We use $url=/v1/chat/completions$ format for batch generations of GPT4o (see Figure~\ref{fig:fsdatagen} for prompt details). The logical solver is Prover 9, a theorem prover suitable to run in python environment with nltk library, that uses CNF conversions, quantifier operations, and skolemization to transform the clauses into a tree format. Parsing errors by Prover 9 occur when the FOL formula cannot be converted to a tree structure because it does not follow specific grammar rules. After filter and parsing stages, we get $10{,}424$ records with FOL statements with a fair distribution of labels. Syntax errors are the errors thrown by the tool and semantic errors are measured by comparing the ground truth label with the solver output.

\paragraph{Quality Evaluation.} We manually evaluated a random sample of 50 records from the dataset (containing $1{,}018$ individual statements) for which Prover9 confirmed correctness of the final outputs. Among these $1{,}018$ translations, we found only 22 errors (approximately $2\%$), half of which were minor and did not affect the semantic interpretation of the FOL statements. We also extracted predicates and performed arity checks and similarity analysis on lemmatized predicate forms across all records, identifying no errors in predicate construction.

\section{In-Context Learning}
\label{app:ICL}
Figure~\ref{fig:fsdatagen} shows the format of the few-shot example for generation of FOLs using data synthesis. It includes an instruction followed by examples which show a sequential generation of predicates, FOL for premise and conclusion. Table~\ref{tab:icldem} reports the detailed accuracy and execution rates for each models and datasets under different number of shots. The \texttt{Qwen-3-4B} model fails to follow the formatting instructions, specifically for ProverQA dataset. 

The prompt for the ICL FOL generation from LMs is presented in Figure~\ref{fig:fol-icl}. The FOL syntax is fixed to represent Prover9 format.

\begin{figure*}[ht]
    \centering
    \begin{tcolorbox}[width=\textwidth, colback=gray!5, colframe=black!50, 
                     title=Few-shot Example with Instruction for Data Generation Process,
                     fonttitle=\bfseries]
    \textbf{Instruction:} You are an expert who works with theorem provers. Given some context and a question, generate the predicates and the first-order logic formula for contexts and question. Here is an example.\\[6pt]
    ---\\
    \textbf{Context:} \\
    Anne is quiet. Erin is furry. Erin is green. Fiona is furry. Fiona is quiet. Fiona is red. Fiona is rough. Fiona is white. Harry is furry. Harry is quiet. Harry is white. Young people are furry. If Anne is quiet then Anne is red. Young, green people are rough. If someone is green then they are white. If someone is furry and quiet then they are white. If someone is young and white then they are rough. All red people are young.\\[6pt]
    \textbf{Question:} \\
    Anne is white.\\[6pt]
    \#\#\#\\
    \textbf{Predicates:}\\
    \quad \texttt{Quiet(x)} $:::$ $x$ is quiet \\
    \quad \texttt{Furry(x)} $:::$ $x$ is furry \\
    \quad \texttt{Green(x)} $:::$ $x$ is green \\
    \quad \texttt{Red(x)} $:::$ $x$ is red \\
    \quad \texttt{Rough(x)} $:::$ $x$ is rough \\
    \quad \texttt{White(x)} $:::$ $x$ is white \\
    \quad \texttt{Young(x)} $:::$ $x$ is young \\[6pt]
    \textbf{Premises:}\\
    \quad \texttt{Quiet(Anne)} $:::$ Anne is quiet.\\
    \quad \texttt{Furry(Erin)} $:::$ Erin is furry.\\
    \quad \texttt{Green(Erin)} $:::$ Erin is green.\\
    \quad \texttt{Furry(Fiona)} $:::$ Fiona is furry.\\
    \quad \texttt{Quiet(Fiona)} $:::$ Fiona is quiet.\\
    \quad \texttt{Red(Fiona)} $:::$ Fiona is red.\\
    \quad \texttt{Rough(Fiona)} $:::$ Fiona is rough.\\
    \quad \texttt{White(Fiona)} $:::$ Fiona is white.\\
    \quad \texttt{Furry(Harry)} $:::$ Harry is furry.\\
    \quad \texttt{Quiet(Harry)} $:::$ Harry is quiet.\\
    \quad \texttt{White(Harry)} $:::$ Harry is white.\\
    \quad $\forall x\,(\texttt{Young}(x) \rightarrow \texttt{Furry}(x))$ $:::$ Young people are furry.\\
    \quad $\texttt{Quiet(Anne)} \rightarrow \neg\texttt{Red(Anne)}$ $:::$ If Anne is quiet then Anne is not red.\\
    \quad $\forall x\,(\texttt{Young}(x) \land \texttt{Green}(x) \rightarrow \texttt{Rough}(x))$ $:::$ Young, green people are rough.\\ 
    \quad $\forall x\,(\texttt{Green}(x) \rightarrow \texttt{White}(x))$ $:::$ If someone is green then they are white.\\
    \quad $\forall x\,(\texttt{Furry}(x)\land \texttt{Quiet}(x) \rightarrow \neg\texttt{White}(x))$ $:::$ If someone is furry and quiet then they are not white.\\
    \quad $\forall x\,(\texttt{Young}(x) \land \texttt{White}(x) \rightarrow \texttt{Rough}(x))$ $:::$ If someone is young and white they are rough.\\
    \quad $\forall x\,(\texttt{Red}(x) \rightarrow \texttt{Young}(x))$ $:::$ All red people are young.\\[6pt]
    \textbf{Conclusion:}\\
    \quad \texttt{White(Anne)} $:::$ Anne is white.\\[6pt]
    ---\\
    \textless\textit{INPUT}\textgreater\\
    \#\#\#\\
    \end{tcolorbox}
    \caption{Few shot example and instruction for FOL generation of FOL dataset using \texttt{gpt4o}.}
    \label{fig:fsdatagen}
\end{figure*}

\begin{figure*}[ht]
    \centering
    \begin{tcolorbox}[
        width=\textwidth, 
        colback=gray!5, 
        colframe=black!50, 
        title=Prompt and a Few-shot Example for ICL FOL Generation, 
        fonttitle=\bfseries
    ]
    \underline{\textbf{System:}} \\
    Given a set of premises and a conclusion, generate the predicates and first-order logic representation of both the premises and the conclusion.\\
    \noindent\rule{\textwidth}{0.4pt}\\
    \underline{\textbf{User:}} \\
    \textbf{Premise:} \\
    Braden advances medical knowledge. If Braden pursues science, then he can either make a groundbreaking discovery or advance medical knowledge, but not both. Braden pursues science. \\
    \textbf{Conclusion:}\\
    Braden makes a groundbreaking discovery.\\
    \underline{\textbf{Assistant:}} \\
    \textbf{{Predicates:}} \\
    AdvancesMedicalKnowledge(x);\\  
        PursuesScience(x);  \\
        MakesDiscovery(x); \\
    \textbf{Premise\_First-order:} \\
    AdvancesMedicalKnowledge(Braden) ::: Braden advances medical knowledge.; \\
    PursuesScience(Braden) $\rightarrow$ ( MakesDiscovery(Braden) $\oplus$ AdvancesMedicalKnowledge(Braden)) ::: If Braden pursues science, then he can either make a groundbreaking discovery or advance medical knowledge, but not both.; \\
    PursuesScience(Braden) ::: Braden pursues science. \\
    \textbf{Conclusion\_First-order:}\\
    MakesDiscovery(Braden) ::: Braden makes a groundbreaking discovery. \\
    \noindent\rule{\textwidth}{0.4pt}\\
    Other Few-Shot Examples\\
    \noindent\rule{\textwidth}{0.4pt}\\
    \underline{\textbf{User:}} \\
    \textbf{Premises:}\\ 
    Lennon has feathers. If Lennon either eats pollen or sips sap (but not both), then he collects nectar. Lennon does not locate food. Lennon does not avoid predators. Any insect that finds shelter either avoids predators or locates food (or both). If Lennon has feathers, then he spreads seeds and flies efficiently. If Lennon hibernates during winter, then he migrates seasonally. Louisa does not avoid predators. Lennon finds shelter or migrates seasonally. If Karim has feathers, then he spreads seeds and flies efficiently. Lennon either collects nectar or spreads seeds, but not both. \\
    \textbf{Conclusion:}\\
    Lennon does not pollinate flowers. \\
    \end{tcolorbox}
    \caption{Prompt for FOL Generation in an ICL setting for Instruct models.}
    \label{fig:fol-icl}
\end{figure*}

\begin{table*}[ht]
\centering
\footnotesize
\begin{tabular}{lllrrrrr}
\toprule
Dataset & Model & Config & Token & Arities & Type & Parsing & Formatting \\
\midrule
ProofWriter & Gemma3 4B Instruct & ICL & 0 & 17 & 7 & 0 & 178 \\
ProofWriter & Gemma3 4B Instruct & SFT\textsubscript{+} & 0 & 1 & 0 & 0 & 6 \\
ProofWriter & Gemma3 4B Instruct & Incremental & 0 & 1 & 0 & 0 & 8 \\
\cdashline{1-8}
ProofWriter & Phi4 mini Instruct & ICL & 0 & 62 & 14 & 0 & 77 \\
ProofWriter & Phi4 mini Instruct & SFT\textsubscript{+} & 0 & 0 & 0 & 0 & 0 \\
ProofWriter & Phi4 mini Instruct & Incremental & 0 & 0 & 0 & 0 & 0 \\
\cdashline{1-8}
ProofWriter & Qwen2.5 3B Instruct & ICL & 8 & 9 & 12 & 0 & 224 \\
ProofWriter & Qwen2.5 3B Instruct & SFT\textsubscript{+} & 0 & 0 & 0 & 0 & 19 \\
ProofWriter & Qwen2.5 3B Instruct & Incremental & 0 & 0 & 0 & 0 & 21 \\
\cdashline{1-8}
ProofWriter & Qwen3 4B Instruct & ICL & 0 & 0 & 0 & 0 & 344 \\
ProofWriter & Qwen3 4B Instruct & SFT\textsubscript{+} & 0 & 0 & 0 & 0 & 30 \\
ProofWriter & Qwen3 4B Instruct & Incremental & 0 & 0 & 0 & 0 & 21 \\

\midrule
FOLIO & Gemma3 4B Instruct & ICL & 1 & 13 & 9 & 1 & 84 \\
FOLIO & Gemma3 4B Instruct & SFT\textsubscript{+} & 0 & 14 & 10 & 1 & 44 \\
FOLIO & Gemma3 4B Instruct & Incremental & 0 & 9 & 12 & 0 & 40 \\
\cdashline{1-8}
FOLIO & Phi4 mini Instruct & ICL & 2 & 28 & 30 & 0 & 56 \\
FOLIO & Phi4 mini Instruct & SFT\textsubscript{+} & 1 & 18 & 13 & 0 & 47 \\
FOLIO & Phi4 mini Instruct & Incremental & 2 & 17 & 18 & 1 & 39 \\
\cdashline{1-8}
FOLIO & Qwen2.5 3B Instruct & ICL & 0 & 3 & 7 & 0 & 148 \\
FOLIO & Qwen2.5 3B Instruct & SFT\textsubscript{+} & 0 & 14 & 6 & 0 & 66 \\
FOLIO & Qwen2.5 3B Instruct & Incremental & 0 & 14 & 6 & 0 & 71 \\
\cdashline{1-8}
FOLIO & Qwen3 4B Instruct & ICL & 0 & 0 & 0 & 2 & 188 \\
FOLIO & Qwen3 4B Instruct & SFT\textsubscript{+} & 0 & 6 & 12 & 2 & 61 \\
FOLIO & Qwen3 4B Instruct & Incremental & 0 & 8 & 11 & 1 & 65 \\

\midrule
ProntoQA & Gemma3 4B Instruct & ICL & 0 & 187 & 127 & 0 & 118 \\
ProntoQA & Gemma3 4B Instruct & SFT\textsubscript{+} & 6 & 106 & 16 & 10 & 103 \\
ProntoQA & Gemma3 4B Instruct & Incremental & 6 & 116 & 11 & 12 & 101 \\
\cdashline{1-8}
ProntoQA & Phi4 mini Instruct & ICL & 0 & 244 & 21 & 0 & 205 \\
ProntoQA & Phi4 mini Instruct & SFT\textsubscript{+} & 0 & 73 & 3 & 0 & 201 \\
ProntoQA & Phi4 mini Instruct & Incremental & 0 & 87 & 2 & 3 & 146 \\
\cdashline{1-8}
ProntoQA & Qwen2.5 3B Instruct & ICL & 6 & 184 & 84 & 0 & 192 \\
ProntoQA & Qwen2.5 3B Instruct & SFT\textsubscript{+} & 0 & 85 & 8 & 1 & 128 \\
ProntoQA & Qwen2.5 3B Instruct & Incremental & 0 & 96 & 5 & 0 & 127 \\
\cdashline{1-8}
ProntoQA & Qwen3 4B Instruct & ICL & 0 & 3 & 0 & 0 & 483 \\
ProntoQA & Qwen3 4B Instruct & SFT\textsubscript{+} & 0 & 32 & 2 & 0 & 155 \\
ProntoQA & Qwen3 4B Instruct & Incremental & 0 & 50 & 1 & 0 & 46 \\

\midrule
ProverQA & Gemma3 4B Instruct & ICL & 95 & 0 & 0 & 0 & 251 \\
ProverQA & Gemma3 4B Instruct & SFT\textsubscript{+} & 4 & 24 & 6 & 0 & 137 \\
ProverQA & Gemma3 4B Instruct & Incremental & 3 & 26 & 5 & 0 & 119 \\
\cdashline{1-8}
ProverQA & Phi4 mini Instruct & ICL & 3 & 1 & 3 & 0 & 257 \\
ProverQA & Phi4 mini Instruct & SFT\textsubscript{+} & 2 & 14 & 2 & 0 & 271 \\
ProverQA & Phi4 mini Instruct & Incremental & 5 & 23 & 3 & 0 & 202 \\
\cdashline{1-8}
ProverQA & Qwen2.5 3B Instruct & ICL & 52 & 0 & 0 & 0 & 447 \\
ProverQA & Qwen2.5 3B Instruct & SFT\textsubscript{+} & 3 & 13 & 1 & 3 & 228 \\
ProverQA & Qwen2.5 3B Instruct & Incremental & 2 & 15 & 2 & 3 & 227 \\
\cdashline{1-8}
ProverQA & Qwen3 4B Instruct & ICL & 0 & 0 & 0 & 0 & 500 \\
ProverQA & Qwen3 4B Instruct & SFT\textsubscript{+} & 1 & 19 & 4 & 0 & 152 \\
ProverQA & Qwen3 4B Instruct & Incremental & 1 & 24 & 5 & 0 & 118 \\

\bottomrule
\end{tabular}
\caption{Error type counts across datasets, models, and methods.}
\label{tab:error_analysis}
\end{table*}

\begin{figure*}[ht]
    \centering
    \begin{tcolorbox}[
        width=\textwidth, 
        colback=gray!5, 
        colframe=black!50, 
        title=Prompt and a Few-shot Example for ICL Predicate Verification, 
        fonttitle=\bfseries
    ]
    \underline{\textbf{System:}} \\
    You are given a premise, a conclusion, and a list of predicates. Check if the predicates correctly represent the premise and conclusion.\\
    Rules:Watch for arity errors: this happens when the same predicate symbol is used with different numbers of arguments.
    Example: Parent(x, y) (2 arguments) vs. Parent(x) (1 argument) → arity error.\\
    If all predicates are correct, output:"correct"\\
    If there are errors, output the corrected list of predicates only\\
    \noindent\rule{\textwidth}{0.4pt}\\
    \underline{\textbf{User:}} \\
    \textbf{Premise:} \\ 
    If a class has prerequisites, the student must take the prerequisites to take the class. If a class has no prerequisites, then the student can take the class CPSC 201 and CPSC 223 are prerequisites for CPSC 323. Intro Microeconomics is the only prerequisite for Intermediate Microeconomics. Intro Geology has no prerequisites. \\
    \textbf{Conclusion:}\\ 
    Intermediate Microeconomics has one prerequisite. \\
    \textbf{{Predicates:}} \\ 
    CanTake(x); CanTake(x, y); Class(x); Prereq(x, y); Student(x); Take(x); Taken(x) \\ 
    \underline{\textbf{Assistant:}} \\
    \textbf{{Predicates:}} \\
    Class(x); Prereq(x, y); Student(x); Take(x, y); CanTake(x, y) \\
    \noindent\rule{\textwidth}{0.4pt}\\
    Other Few-Shot Examples\\
    \noindent\rule{\textwidth}{0.4pt}\\
    \underline{\textbf{User:}} \\
    \textbf{Premises:}\\ 
    Yale University is a private Ivy League research university. Yale University moved to New Haven in 1716. Yale university's endowment was valued at 42.3 billion. A list of residential colleges at Yale: Benjamin Franklin College, Berkeley College, Branford College, Davenport College, Ezra Stiles College, Grace Hopper College, Jonathan Edwards College, Morse College, Pauli Murray College, Pierson College, Saybrook College, Silliman College, Timothy Dwight College, and Trumbull College. \\
    \textbf{Conclusion:}\\
    Yale University has the largest university endowment of any educational institution. \\
    \textbf{{Predicates:}} \\
    Is(x, y); Endowment(x, y); ResidentialCollege(x); ResearchUniversity(x); IvyLeague(x); MoveTo(x, y); List(x, y); Value(x, y) \\
    \end{tcolorbox}
    \caption{Prompt for Predicate Verification in an ICL setting for Gemma4B Instruct model.}
    \label{fig:pred-icl}
\end{figure*}

\section{Error Distribution in FOL Generations}
\label{app:errordist}
Table~\ref{tab:error_analysis} reports the statistics of syntax errors in detail. Smaller LMs are more prone to formatting errors, as they often get stuck in a loop of greedy decoding—repeatedly choosing the same next prediction—which makes the translation invalid. After fine-tuning the LMs, we observe a significant reduction in these formatting errors.

\section{Verification with LM}
\label{app:verifier}
We apply a simple plug-and-play verification module on top of incremental inference using 3-shot examples. One example demonstrates the `correct’ value, while the other two illustrate different forms of arity errors. These include perturbed arity forms, where an additional variable is inserted into a predicate, with the target correction being the corresponding single predicate. For FOLIO, we use examples generated by passing the training data to an LLM, extracting and reusing specifically the arity errors. For ProntoQA, which is a test-only dataset, we use data from ProofWriter. For ProverQA, arity errors are manually constructed. An example prompt is shown in Figure~\ref{fig:pred-icl}.

\section{Further Details of SFT and Inference}
\label{appE}
The models are fine-tuned for three epochs with a fixed batch size of 64, a learning rate of $1\times10^{-4}$, a weight decay of 0.01, and a LoRA configuration~\cite{hu2022lora} adapted to the language model on a single A100 GPU. Training uses a causal language modeling loss applied to inputs formatted as \texttt{\{NL \textbackslash n\textbackslash n Predicates; FOL\}}, with gradients computed over the entire input.   
 
Inference is performed using 8-bit quantization with the trained LoRA adapter, a temperature of 0.1, and a variable \texttt{maximum\_new\_tokens} value depending on the dataset. \textsc{ProofWriter} and \textsc{ProverQA} require a larger number of tokens during inference compared to \textsc{FOLIO} and \textsc{ProntoQA} due to their longer input sizes.

\section{Further details on Language Models}
\label{appH}
We use four instruction-tuned language models from different model families. All model weights are downloaded from Hugging Face\footnote{\url{https://huggingface.co}}: \texttt{Qwen/Qwen2.5-3B-Instruct}, \texttt{Qwen/Qwen3-4B-Instruct-2507}, \texttt{google/gemma-3-4b-it}, and \texttt{microsoft/Phi-4-mini-instruct}. The \textsc{Qwen} models have 3B and 4B parameters, while \textsc{Gemma} and \textsc{Phi} each have 4B parameters.

\end{document}